\documentclass{article}

\PassOptionsToPackage{numbers, compress}{natbib}
\usepackage{subcaption}
\usepackage{wrapfig}
\usepackage[main, final]{neurips_2025}
\usepackage{graphicx}

\usepackage{amsmath,amsfonts,bm}

\def\eqref#1{equation~\ref{#1}}

\def\1{\bm{1}}

\def\vx{{\bm{x}}}

\DeclareMathAlphabet{\mathsfit}{\encodingdefault}{\sfdefault}{m}{sl}
\SetMathAlphabet{\mathsfit}{bold}{\encodingdefault}{\sfdefault}{bx}{n}

\def\gP{{\mathcal{P}}}

\usepackage[table]{xcolor}
\usepackage{enumitem}

\usepackage[utf8]{inputenc} %
\usepackage[T1]{fontenc}    %
\usepackage{hyperref}       %
\usepackage{url}            %
\usepackage{booktabs}       %
\usepackage{amsfonts}       %
\usepackage{nicefrac}       %
\usepackage{microtype}      %
\usepackage[usenames,dvipsnames]{xcolor}         %
\usepackage{multirow}
\usepackage{colortbl}
\usepackage{siunitx}
\usepackage{cleveref}
\usepackage[most]{tcolorbox}
\newtcolorbox{myblock}[1][]{
    colback=gray!10!white, 
    colframe=gray!80!black,        
    fonttitle=\bfseries,        %
    title=#1,                   
    boxrule=0.5mm,              %
    rounded corners,            %
    before skip=10pt,           %
    after skip=10pt,            %
}

\usepackage{scalerel}
\def\smileyface{\scalerel*{\includegraphics{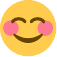}}{\textrm{\textbigcircle}}}

\title{URLs Help, Topics Guide: \\
Understanding Metadata Utility in LLM Training}

\author{%
 Dongyang Fan\quad Vinko Sabolčec\quad Martin Jaggi\\
 EPFL, Switzerland \\
  \texttt{firstname.lastname@epfl.ch} \\
}

\begin{document}

\maketitle

\begin{abstract}
Large Language Models (LLMs) are commonly pretrained on vast corpora of text without utilizing contextual metadata such as source, quality, or topic, leading to a context-free learning paradigm. While recent studies suggest that adding metadata like URL information as context (i.e., auxiliary inputs not used in the loss calculation) can improve training efficiency and downstream performance, they offer limited understanding of which types of metadata are truly effective and under what conditions. In this work, we conduct a systematic evaluation and  find that not all metadata types contribute equally. Only URL context speeds up training, whereas quality scores and topic/format domain information offer no clear benefit. Furthermore, the improved downstream performances of URL conditioning emerge only when longer prompts are used at inference time. In addition, we demonstrate that context-aware pretraining enables more controllable generation than context-free pretraining, in a classifier-free guidance fashion. Although topic and format metadata do not accelerate training, they are effective for steering outputs, offering human-interpretable control over generation. Our codes are available at \href{https://github.com/fan1dy/metadata-enhanced-pretrain-datapipeline}{this link}.

\end{abstract}

\section{Introduction}
Large Language Models (LLMs) have become increasingly integrated into a wide range of real-world applications, prompting both researchers and practitioners to explore optimal strategies for their development. A critical component of this process is the pretraining phase, which serves as the foundational stage in which models acquire general-purpose linguistic and world knowledge from vast corpora of text. Even with a simple next-token prediction loss, the scale of data and model size enables emergent generalization of knowledge~\cite{du2024understanding}.

Conventional LLM pretraining is typically \emph{context-free}, in the sense that training data consists solely of raw textual inputs, without the inclusion of any associated metadata or external contextual signals. Meta information, such as document source, author identity, timestamps, or topic tags—although potentially valuable—are systematically discarded during preprocessing. As a result, the model learns to predict the next token based on the intrinsic structure and statistical patterns present within the text itself, without leveraging any auxiliary semantic or structural cues. A natural research question arises: Can the pretraining be improved if contextual information is included?

While prior studies~\citep{allenzhu2024physicslanguagemodels33,gao2025metadataconditioningaccelerateslanguage} suggest that incorporating URLs as context (auxiliary inputs not used in the loss calculation) during training can accelerate pretraining and enhance downstream performance, recent findings by \citet{higuchi2025doesmetadataconditioningnot} caution that this is not always the case. Existing work lacks a comprehensive understanding of \emph{which types of metadata are most impactful and under what conditions they offer the greatest benefit}. Our study aims to fill this gap, providing empirical evidence based on real-world experiments.

Fundamentally, context-aware pretraining endowed LMs with the ability to perform conditional autoregressive generation. This conditional generation capability, developed through exposure to metadata, can potentially be leveraged to steer the model's behavior using a classifier-free guidance approach~\citep{ho2022classifierfreediffusionguidance}, i.e., the distinction between context-conditioned and context-free generation can be further amplified to highlight the influence of contextual information. We aim to investigate \emph{whether context-aware pretraining can be utilized for more controllable generation than context-free pretraining. }

In response to the previously raised questions, our findings and contributions can be summarized as follows:
\begin{itemize}
    \item We conduct a comprehensive study on context-conditioned LLM pretraining, highlighting its potential benefits and limitations at both pretraining and inference stages (Section~\ref{sec: methods}).

    \item We examine different types of metadata and find that only URL as context can speed up pretraining (Section~\ref{sec: only-URL-conditioning-works}). When it comes to downstream evaluation, URL-conditioned pretraining only enhances the performance with longer prompts (Section~\ref{sec: longer-prompts}). 
    \item We demonstrate that the metadata-conditioned pretrained model is more steerable than the standard pretrained model. While topic and format domain information does not accelerate pretraining, it offers effective and human-interpretable steering (Section~\ref{sec: context-aware generation}). 
\end{itemize}

\section{Related Work}
\paragraph{Metadata for LLM pretraining.} \citet{allenzhu2024physicslanguagemodels33} demonstrate through controlled synthetic experiments that adding a special token to high-credibility data can be beneficial. They further hypothesize that appending domain names (e.g., wikipedia.org) to each document in the pretraining corpus would yield similar gains. This hypothesis is experimentally validated by \citet{gao2025metadataconditioningaccelerateslanguage}, who report a 33\% speedup in pretraining simply by prepending URL domains. However, to ensure the model remains effective on standard, domain-free text, they introduce a "cool-down" phase using non-contextual data. Following this, \citet{zhu2025powercontextenhancedlearningllms} formally proves that context-enhanced learning can be exponentially more sample-efficient when the model is capable of in-context learning, that is, it can effectively utilize contextual information presented during inference to perform tasks. More recently, \citet{higuchi2025doesmetadataconditioningnot} caution that metadata conditioning is not universally effective. Buidling up on the experimental setup of \cite{allenzhu2024physicslanguagemodels33}, they found that metadata conditioning only helps when the input prompt is sufficiently long to reveal latent semantics. With shorter prompts, it can even degrade performance. Most prior studies conduct experiments solely on synthetic datasets, where the metadata consists of the datasets' production rules. However, in real-world scenarios, metadata cannot fully capture the complexity of natural language. Therefore, we aim to explore the extent to which such metadata can actually be helpful.

\paragraph{Guidance in LLMs.} In LLMs, next-token generation is essentially a sampling process from a probability distribution over the entire vocabulary. This makes it possible to steer the generation by modifying the token probability distribution. \citet{liu-etal-2021-dexperts} guide the generation of a base LLM by leveraging the difference between the outputs of an expert model and an anti-expert model. \citet{li-etal-2023-contrastive} adjust the sampling distribution by down-weighting tokens favored by an amateur model. \citet{sanchez2023staytopicclassifierfreeguidance} introduced the idea of classifier-free guidance (CFG)—originally used in text-to-image generation—to text-only generation. Building on this, \citet{he2025contextsteeringcontrollablepersonalization} applied CFG to steer LLMs for personalization purposes, showing potential for controllable personalization. However, all of these approaches rely on context-free pretrained LLMs. In this work, we investigate can
context-conditioned pretrained LLMs enable more effective context steering. 

\section{Method}
\label{sec: methods}
\begin{figure}
    \centering
    \includegraphics[width=\linewidth]{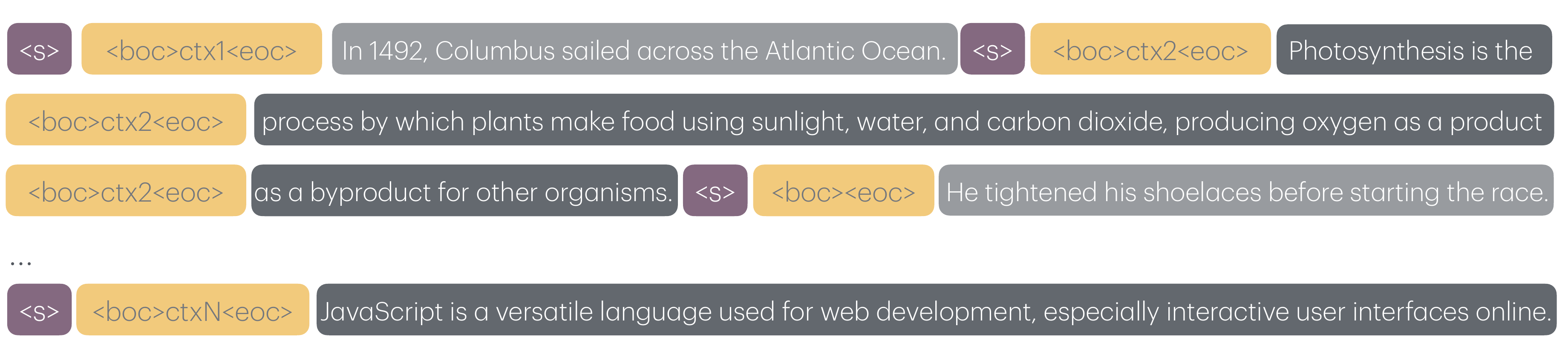}
    \caption{An example of our context-aware tokenization. Each document begins with a default beginning-of-sequence (\texttt{<s>}) token. For each sequence, a context segment wrapped in beginning-of-context (\texttt{<boc>}) and end-of-context (\texttt{<eoc>}) is inserted after \texttt{<s>} and before the main text. If a document is too long and split into multiple sequences, the context is prepended to each one. Although the context is added to every sequence, it may be empty. In 90\% of the corpus, we include a non-empty context; in the remaining 10\%, the context is left empty. Contexts can be URL, quality score, or topic/format domains depending on user choices.}
    \label{fig:tokenization-diagram}
    \vspace{-1em}
\end{figure}

\subsection{Context-Conditioned Pretraining}
To properly organize the contextual information, we introduce two new tokens: \texttt{<|context\_begin|>} and \texttt{<|context\_end|>}\footnote{for illustration purposes, we use \texttt{<boc>} and \texttt{<eoc>} to represent these two tokens}. The meta information is inserted between the two tokens, and prepended before the document contents. During training time, we mask out the loss calculation for the prepended contexts, so the loss is solely calculated over the standard texts, making it comparable to standard context-free pretraining. To avoid confusion, we use \emph{context} to refer to the auxiliary meta information not used in the
loss calculation during training time, and use \emph{prompts} for the texts given to the pretrained model for generation.

\begin{wrapfigure}{rt}{0.6\textwidth}
    \centering
    \includegraphics[width=\linewidth]{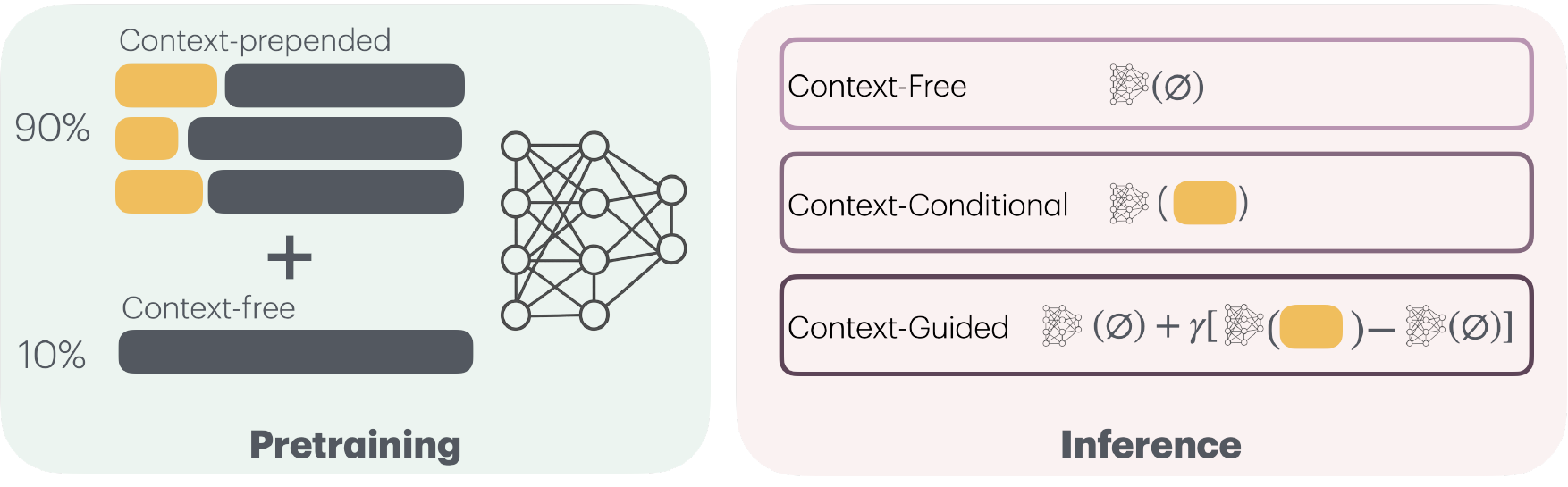}
    \caption{Diagram of our two-stage investigation. During pretraining time, we feed a uniform mixture of 90\% context-prepended texts and 10\% context-free standard texts into the model. During inference time, we compare three different generation sampling methods.}
    \label{fig:enter-label}
    \vspace{-1em}
\end{wrapfigure}

Context is introduced at the sequence level and is specific to each document. To ensure the model remains capable of handling context-free text at inference time, we uniformly interleave context-prepended and context-free documents during pretraining, using a 90\%:10\% ratio. The context enhanced corpus is illustrated in Figure~\ref{fig:tokenization-diagram}, where one document can be split into several sequences and contexts are prepended to each of them. Unlike the MeCo approach~\cite{gao2025metadataconditioningaccelerateslanguage}, which applies 90\% context-prepended training followed by a 10\% context-free cooldown phase, our uniform mixture strategy allows any intermediate checkpoint to be immediately used for context-free inference.

\subsection{Context-Aware Generation}

At inference time, leveraging the conditioning capabilities learned during pretraining, we can guide the LLM to generate text based on a given context. Next-token prediction involves sampling from a probability distribution ($\gP^{\text{LLM}}$) over the entire vocabulary. To incorporate contextual signals, we modify this distribution accordingly. We define the three following sampling methods for generation:
\begin{enumerate}[label=\arabic*)]
    \item Context-Free sampling: this is the standard sampling given by an LLM, conditioning on empty contexts.
    \begin{equation}
   x^t \sim   \gP^{\text{LLM}}(x|\vx^{1:t-1}, \emptyset) \tag{ctx-free}
\end{equation}
\item Context-Conditioned sampling: for each generation, we prepend \texttt{<boc>ctx<eoc>} to the prompt.
\begin{equation}
   x^t \sim   \gP^{\text{LLM}}(x|\vx^{1:t-1}, \text{ctx}) \tag{ctx-conditioned}
\end{equation}
\item Context-Guided sampling: following the classifier-free guidance in diffusion models~\cite{ho2022classifierfreediffusionguidance}, we amplify the impact of the context with the following sampling probability:
\begin{equation}
    x^t \sim \gP^{\text{LLM}}(x|\vx^{1:t-1},\emptyset) \cdot   \left( \frac{\gP^{\text{LLM}}( x|\vx^{1:t-1},\text{ctx})}{\gP^{\text{LLM}}(x|\vx^{1:t-1},\emptyset)}\right)^\gamma \tag{ctx-guided}
 \end{equation}
 During inference, in practice, we modify the generation in the logit ($\Pi$) level, that is
\begin{equation}
\label{eq: modified-logits}
    x^t \sim \Pi^{\text{LLM}}(x|\vx^{1:t-1},\emptyset) + \gamma   \left( \Pi^{\text{LLM}}(x|\vx^{1:t-1},\text{ctx})-\Pi^{\text{LLM}}(x|\vx^{1:t-1},\emptyset)\right)
 \end{equation}
When $\gamma =0$, this coincides with context-free generation and when $\gamma=1$, this coincides with context-conditioned generation. With $\gamma>1$, we amplify the guidance of the context. 
\end{enumerate}

It is worth noting that the softmax output after context-guidance is no longer a faithful\footnote{By faithful, we mean a probability distribution that can be trusted as a true representation of uncertainty, without distortion, manipulation, or miscalibration.} probability distribution, while context-free and context-conditioned generation still follow a faithful probability distribution.

\section{Experiments}
\label{sec: experiments}
\textbf{Model.} We adopt the Llama model architecture~\citep{grattafiori2024llama3herdmodels} with 16 layers,
a hidden size of 2048, a sequence length of 4096, and a batch size of 504 (resulting in 2.06 million tokens). The model has 1.5 billion parameters. We follow the Cosine learning schedule, applying 2000 warmup
steps. AdamW optimizer is used with regularization strength 0.1~\citep{loshchilov2019decoupledweightdecayregularization}. A max learning rate of 3e-4 is applied and cools down to learning rate
3e-5 at the end of training. To train the models, we use the Megatron-LM framework~\cite{narayanan2021efficientlargescalelanguagemodel}.

\textbf{Dataset.} We use FineWeb-Edu dataset~\cite{lozhkov2024fineweb-edu}, which is a high-quality English-only dataset. The dataset has a lot of meta data available, such as URL source, quality score and token counts per document. We sub-sample FineWeb-Edu and tokenize it using Nemo tokenizer\footnote{https://mistral.ai/news/mistral-nemo}, with two additionally introduced tokens. Throughout training, we randomly sample 100B tokens.

\textbf{Evaluation Benchmarks}. As standard practice, we evaluate models on general knowledge understanding using LM-Eval-Harness developed by \citet{eval-harness}. The benchmarks used are Arc-Easy~\citep{clark2018think}, Arc-Challenge~\citep{clark2018think}, CommonSense QA (CSQA, \citealp{talmor2018commonsenseqa}), MMLU~\citep{hendrycks2020measuring}, PIQA~\citep{bisk2020piqa}, Social IQA (SIQA, \citealp{sap2019socialiqa}), HellaSwag (HS, \citealp{zellers2019hellaswag}), Lambada (LBD, \citealp{paperno2016lambada}) and Winogrande (WG, \citealp{sakaguchi2021winogrande}).

We compare the following runs to gain a full insight over how  can contexts potentially enhance LLM pretraining. 
\begin{enumerate}[itemsep=0pt]
\item Baselines

\begin{itemize}[itemsep=0pt]
    \item \emph{Standard Pretraining}: Each document is simply prepended with a begin-of-sequence token <s>.
    \item \emph{Positional Token Ablation}: Standard pretraining where each sequence is prepended with <boc><eoc> tokens. This setup tests whether introducing a positional token at the start of each sequence alone has a beneficial effect.
    \item \emph{MeCo}: A two-phase approach proposed by \citet{gao2025metadataconditioningaccelerateslanguage}—90\% of training is URL-conditioned, followed by a 10\% cooldown phase using standard (unconditioned) data.
\end{itemize}

\item Types of Metadata
\begin{itemize}[itemsep=0pt]
 \item \emph{Full URL (URL)}: The original source from which the document was crawled.

 \item \emph{Quality Score (QS)}: Provided by the FineWeb-Edu dataset, these scores are generated by a classifier trained on 450,000 web samples, each annotated by LLaMA 3 with a score from 0 (not educational) to 5 (highly educational), reflecting the sample's educational value.

 \item \emph{Domain Information (DI)}: Each document is labeled with topic and format domains derived from WebOrganizer~\cite{wettig2025organizewebconstructingdomains}, where topic and format domains are returned by pretrained classifiers. There are 24 categories per taxonomy. In total, we have 576 different Domain Information types.
 \end{itemize}

 \item Combinations of Metadata
\begin{itemize}[itemsep=0pt]
\item \emph{Full URL + Quality Score}: Combines the document’s source with its educational quality.

\item \emph{Full URL + Domain Information}: Combines the source with topic and format domain labels.

\item \emph{Full URL + Quality Score + Domain Information}: Incorporates all available metadata to condition the model on source, quality, and content structure.
\end{itemize}

\end{enumerate}

\subsection{Not All Metadata Conditioning Speed-up Pretraining}
\label{sec: pre-training-speed-up}

\subsubsection{Training Speed-up with URL Conditioning} 
\label{sec: only-URL-conditioning-works}

\begin{figure*}[t!]
    \centering
    \begin{subfigure}[t]{0.4\textwidth}
        \centering
\includegraphics[width=\linewidth]{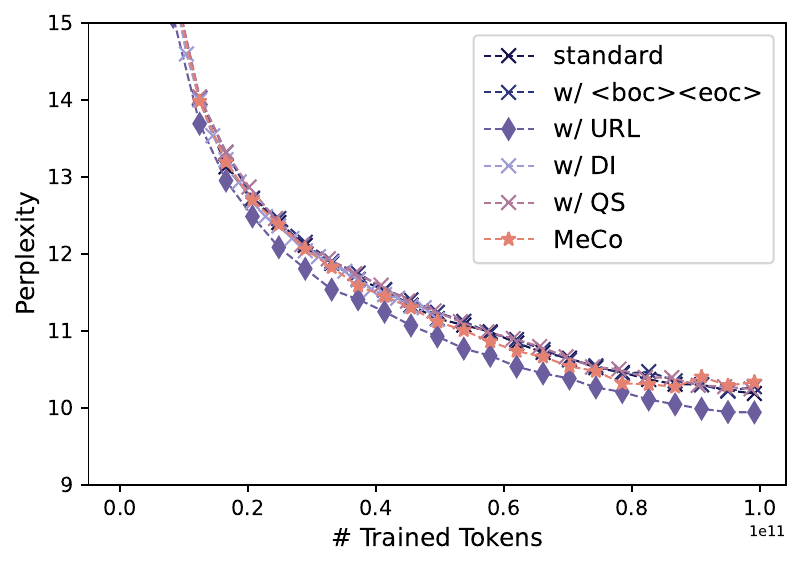}
    \end{subfigure}%
    ~ 
    \begin{subfigure}[t]{0.4\textwidth}
        \centering
        \includegraphics[width=\linewidth]{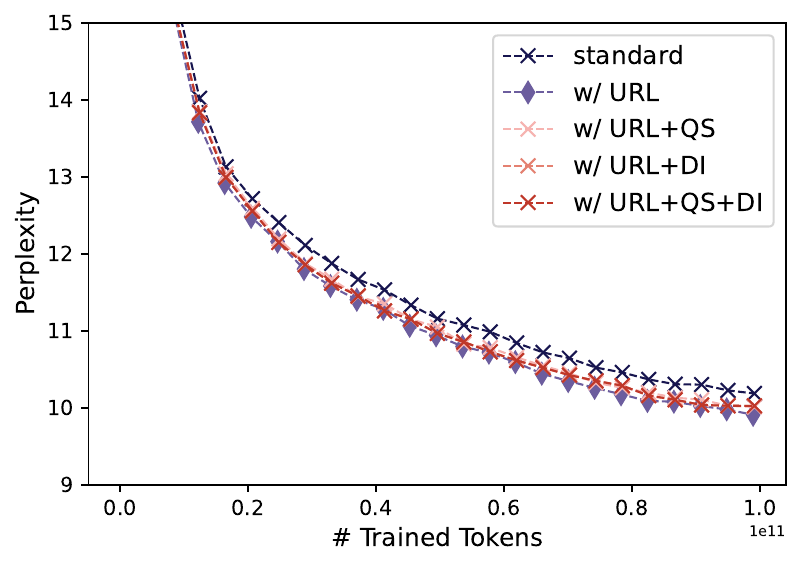}
    \end{subfigure}
    \caption{Training perplexity versus the amount of consumed tokens. Prepending the URL leads to a faster decrease in perplexity.} \label{fig:training-loss-comparison}
    \vspace{-1em}
\end{figure*}

We begin by investigating whether metadata can facilitate training, specifically, whether pretraining can be accelerated when additional contextual information aids knowledge acquisition. The left panel of Figure~\ref{fig:training-loss-comparison} presents the training perplexity across runs with different metadata types. The two additional tokens introduced have minimal effect on learning dynamics, indicating that any observed differences stem from the contextual content itself. Among the three types of context evaluated, only the URL consistently results in lower perplexity from the very beginning of training.

\textbf{Is the improved performance stemming from longer contexts?} Given that URLs are substantially longer than the other two metadata types, one might wonder whether their effectiveness stems merely from increased context length. To examine this, we combined the other metadata types with URLs to see if additional context would further improve training. From the right panel of Figure~\ref{fig:training-loss-comparison}, the answer is No. This indicates that it is not the increased context length that speeds up the training, but the context content itself.

\begin{figure}
    \centering    \includegraphics[width=\linewidth]{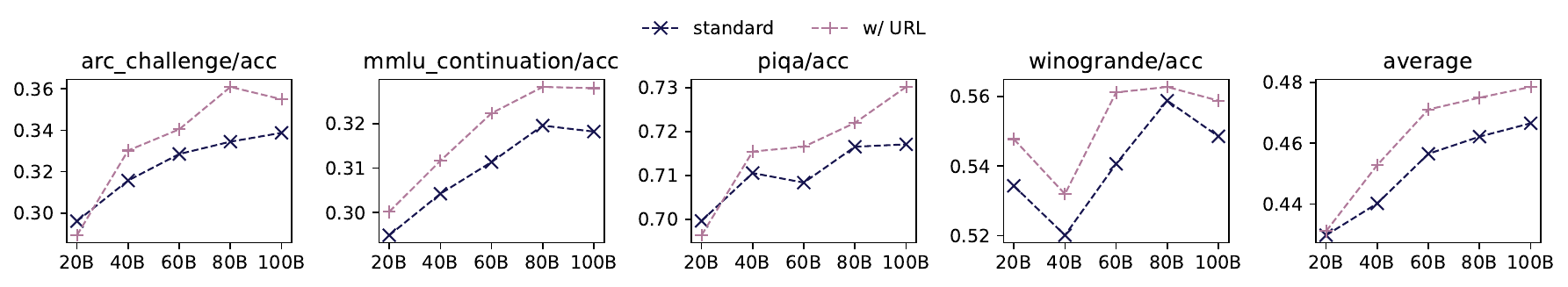}
    \caption{URL-conditioned pretraining achieves the same downstream evaluation performances of 100B-token standard pretraining  with only 60B tokens. The same plots with respect to all tasks are provided in Figure~\ref{fig:eval-speed-up-02}. }
    \label{fig:training-speed-up-eval}
\end{figure}

\textbf{Does the acceleration translate into downstream performance as well?}
With five-shot evaluation, we see better performance for the URL-conditioned model in every task. For the 9-task average, the URL-conditioned model can match the performance of the standard pretrained model on 100B tokens with only 60B tokens, achieving a 40\% significant acceleration, as shown in Figure~\ref{fig:training-speed-up-eval}. The acceleration in terms of downstream performance is greater than training loss. 

\begin{figure*}[t!]
    \centering
    \begin{subfigure}[t]{0.32\textwidth}
        \centering
\includegraphics[width=\linewidth]{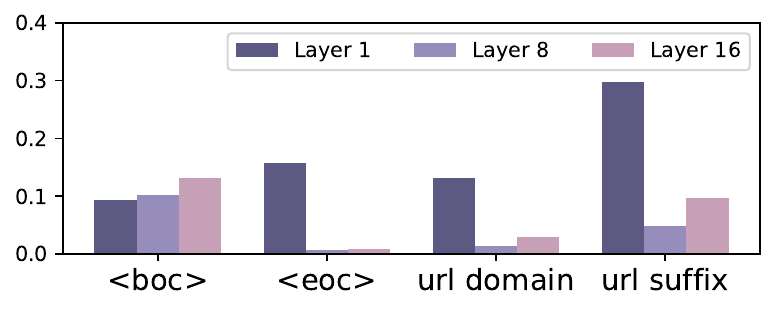}
    \end{subfigure}%
    ~ 
    \begin{subfigure}[t]{0.32\textwidth}
        \centering
        \includegraphics[width=\linewidth]{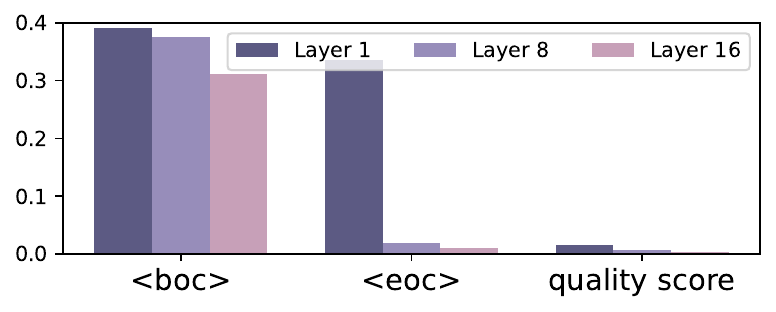}
    \end{subfigure}
    ~
        \begin{subfigure}[t]{0.32\textwidth}
        \centering
        \includegraphics[width=\linewidth]{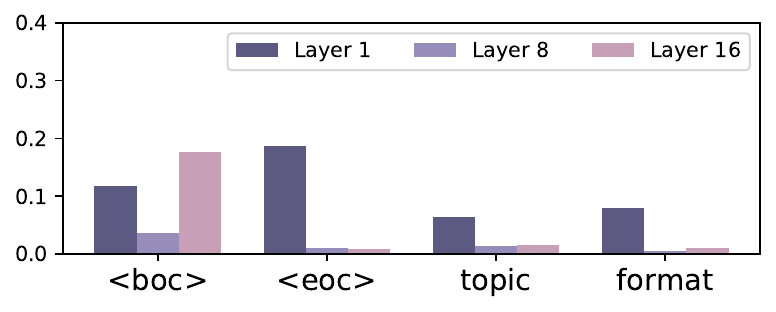}
    \end{subfigure}
    \caption{Average attention to different parts of different prepended contexts. Take \texttt{https://en.wikipedia.org/wiki/Metadata\#Standards} as an example. \texttt{en.wikipedia.org} is the URL domain and \texttt{/wiki/Metadata\#Standards} is the URL suffix. More details in Appendix~\ref{app: attention-pattern}.} \label{fig: average-attention-pattern-3-subplots}
    \vspace{-1em}
\end{figure*}

\textbf{Why only URL helps?} We have already controlled for the confounding factor — the length of metadata. However, it remains unclear why only the URL contributes meaningfully. To gain insight into this, we visualized the attention patterns of the document texts toward different parts of their corresponding contexts in Figure~\ref{fig: average-attention-pattern-3-subplots}. Except for the URL-conditioned model, the other models fail to allocate substantial attention to the informative parts of the contexts in the early layers. We \emph{hypothesize} that the early-layer attention to meaningful URL components may play a key role in the effectiveness of URL prepending, even if later layers focus on less informative tokens. Moreover, it is important to recognize that URLs inherently encode both domain information and document quality, which may differ from human-assigned DI and QS labels. This distinction likely also contributes to the superior performance of the URL feature.

\begin{myblock}
\textbf{Takeaway 1.}
Only URL conditioning has a positive effect in speeding up pretraining; conditioning on quality scores or topic/format domains yields no noticeable change.
\end{myblock}

\subsubsection{Downstream Performance Gains from Longer Prompts}
\label{sec: longer-prompts}

\textbf{When does URL conditioning work?} To better understand the impact of context conditioning, we perform a systematic evaluation across downstream tasks by comparing zero-shot and five-shot performance. We notice a very interesting phenomenon: \emph{The URL-conditioned pretrained model shows increased downstream evaluation performance with five-shot evaluation, but not zero-shot}. For the rest, we do not see substantial improvement over standard pretraining. Our results confirm recent findings from \citet{higuchi2025doesmetadataconditioningnot}, where longer prompts are needed to infer latent semantics and thus to help evaluation.

\textbf{Conflicting signals of different metadata types.} Interestingly, the benefits of URL conditioning are negated when additional metadata is introduced as context. For example, combining quality scores with URL conditioning appears to introduce conflicting signals, leading to downstream performance that is even lower than that of the standard pretrained model. This suggests that the latent clusters inferred by the language model from the URLs may differ from the predefined clusters we assigned. Notably, we did not manage to reproduce the results of the MeCo baseline. Possible reasons could be different datasets and models, and that LM-Eval-Harness~\cite{eval-harness} is used as our evaluation framework is used in our experiments, while OLMES~\citep{gu-etal-2025-olmes} was used in the MeCo paper. Nonetheless, we still observe the speeding-up effect of URL prepending.

\begin{table}[h!]
\vspace{-1em}
\caption{0shot evaluation results. When evaluating, context-free generation is used, i.e., only \texttt{<boc><eoc>} tokens are prepended and the context string is empty.}
\centering
\resizebox{.9\textwidth}{!}{
\begin{tabular}{l c c c c c c c c c c c}
\toprule
  &\textbf{Arc-C} & \textbf{Arc-E} & \textbf{CSQA} &  \textbf{MMLU} & \textbf{PIQA} & \textbf{SIQA} & \textbf{HS}  & \textbf{LBD} & \textbf{WG} & \textbf{Avg} \\
\midrule
standard & 32.9 & 68.9	&42.3	&31.2&	71.3&	40.6	&42.6&	33.0	&57.1&	46.7 \\
+ \texttt{<boc><eoc>} & 33.1	&68.4&	41.8	&32.0	&72.1	&40.2&	42.4&	33.4	&56.6&	46.7 \\
\arrayrulecolor{black!30} \midrule 
 {+ \texttt{<boc>} URL \texttt{<eoc>}}  & 33.5 & 69.8 &	41.6	&32.0	&73.4	&41.3&	42.8&	33.2	&54.4	& \textbf{46.9}\\
 {+ \texttt{<boc>} QS \texttt{<eoc>}} & 30.9	&68.4	&38.3	&30.9&	72.4	&41.0	&42.0	&34.4&	53.5&	45.8  \\
 {+ \texttt{<boc>} DI \texttt{<eoc>}}& 32.1& 	68.4	& 41.9	& 31.8	& 71.3& 	41.1& 	42.1	& 34.2& 	53.4& 	46.3  \\
 \arrayrulecolor{black!30} \midrule 
  + \texttt{<boc>} {URL, QS} \texttt{<eoc>}& 31.2& 	68.7	& 38.2& 	31.7	& 71.3	& 40.1	&42.2	&34.6	&54.2	&45.8\\
 + \texttt{<boc>} URL, QI \texttt{<eoc>}  & 31.2& 	68.1& 	37.8	& 31.8& 	71.9& 	40.6	& 42.0	& 33.7	& 56.5	& 46.0\\
 {+ \texttt{<boc>} URL, QS, DI \texttt{<eoc>}} & 32.4	& 69.4	& 40.8	& 31.8	& 71.8	& 39.8& 	42.0& 	33.0	& 54.9& 	46.2 \\
  \arrayrulecolor{black!100} \midrule 
 MeCo & 33.0	 &68.9	 &39.7	 &31.8	 &71.3	 &40.1	 &41.9	 &33.0	 &55.6	 &46.2\\

\bottomrule
\end{tabular}
}
\vspace{-1em}
\end{table}

\begin{table}[h!]
\caption{5shot evaluation results. When evaluating, context-free generation is used, i.e., only \texttt{<boc><eoc>} tokens are prepended and the context string is empty.}
\centering
\resizebox{.9\textwidth}{!}{
\begin{tabular}{l c c c c c c c c c c c}
\toprule
  & \textbf{Arc-C} & \textbf{Arc-E} & \textbf{CSQA} &  \textbf{MMLU} & \textbf{PIQA} & \textbf{SIQA} & \textbf{HS}  & \textbf{LBD} & \textbf{WG} & \textbf{Avg} \\
\arrayrulecolor{black!30} \midrule
standard & 33.9	& 68.6	& 45.1	& 31.8	& 71.7	& 41.5 &	41.9 &	30.6	& 54.9	& 46.7 \\
+ \texttt{<boc><eoc>} & 34.0	& 69.2	&43.5	&31.9	&71.9	&41.5	&42.5	&30.8	&55.2	&46.7\\
\arrayrulecolor{black!30} \midrule

 {+ \texttt{<boc>} URL \texttt{<eoc>}} & 35.5	&71.7	&46.8	&32.8	&73.0	&41.7&	42.5	&30.8	&55.9&	\textbf{47.8}\\
{+ \texttt{<boc>} QS \texttt{<eoc>}} & 32.7	& 69.6	& 41.0	& 31.8	& 72.1& 	41.6	& 43.0& 	32.4& 	55.2	& 46.6 \\

 +  \texttt{<boc>} DI \texttt{<eoc>}& 34.0	& 69.8&	42.6	&32.1&	71.9	&40.3	&42.0&	31.8	&55.5&	46.7\\
 \midrule
 {+  \texttt{<boc>} URL, QS}\texttt{<eoc>} &32.5&	67.8	&42.8&	31.6&	71.4	&41.7	&41.8&	32.0&	53.0	&46.1 \\
  +  \texttt{<boc>} URL, DI \texttt{<eoc>} & 33.9	&67.4	&42.2	&31.4	&71.8&	41.5	&42.3	&32.9	&57.0&	46.7\\
 {+ \texttt{<boc>} URL, QS, DI \texttt{<eoc>}} & 32.9&	68.6	&44.6	&31.9&	71.6	&39.8	&41.9	&31.8&	51.0&	46.0\\
 \arrayrulecolor{black!100} \midrule
 MeCo & 32.6 & 69.7 & 46.3 & 32.5 & 72.0 & 39.9 & 41.6 & 31.3 & 54.0 & 46.7 \\
\bottomrule
\end{tabular}
}
\label{tab: 5shot-eval}
\vspace{-1em}
\end{table}

\begin{myblock}
\textbf{Takeaway 2.} URL-conditioned pretraining only benefits evaluation for longer prompts, for example, 5-shot. There is no noticeable distinction for 0-shot evaluation.
    
\end{myblock}

\subsection{Context-Aware Generation}
\label{sec: context-aware generation}

\subsubsection{Context-Conditioned Generation}

The results from Table~\ref{tab: 5shot-eval} are from context-free generation using a context-conditioned model. We now test whether the performance can further be improved if correct contexts are given at test time. For each task, we conditioned the generation on an contexts that match the task content. We manually select these contexts. For Quality Score context, we simply prepend \texttt{Quality Score: 5} in the inference time. The contexts we used are presented in Table~\ref{contexts-for-inference} in the Appendix. Conditioning on provided contexts, we observed increased downstream performance, especially in URL- and DI-conditioned pretrained models.

\begin{table}[h!]
\caption{Enhanced downstream task performance with context-conditioned generation compared to context-free generation. Three different contexts are tested here, that is, URL, Domain Information (DI) and Quality Score (QS).}
\centering
\resizebox{\textwidth}{!}{
\begin{tabular}{l l c c c c c c c c c c c}
\toprule
 \textbf{Model} & \textbf{Context} & \textbf{Arc-C} & \textbf{Arc-E} & \textbf{CSQA} &  \textbf{MMLU} & \textbf{PIQA} & \textbf{SIQA} & \textbf{HS}  & \textbf{LBD} & \textbf{WG} & \textbf{Avg} \\
  \toprule
\multirow{2}{*}{URL-conditioned} & w/o  context & 35.5	&71.7	&46.8	&32.8	&73.0	&41.7&	42.5	&30.8	&55.9&	47.8\\
& w/ URL context & 35.5 & 	72.2	 & 47.7	 & 32.9	 & 72.3	 & 43.0	 & 43.0	 & 32.5	 & 56.0	 & 48.3 $\uparrow$
\\
\arrayrulecolor{black!30} \midrule
\multirow{2}{*}{DI-conditioned} & w/o context & 34.0	& 69.8&	42.6	&32.1&	71.9	&40.3	&42.0&	31.8	&55.5&	46.7 \\

& w/ DI context & 33.9	&70.9	&46.0	&32.3	&72.4&	41.7	&42.2	&33.2	&55.1&47.5 $\uparrow$ \\
\arrayrulecolor{black!30} \midrule
\multirow{2}{*}{QS-conditioned} & w/o  context &  32.7	& 69.6	& 41.0	& 31.8	& 72.1& 	41.6	& 43.0& 	32.4& 	55.2	& 46.6\\

& w/ QS context & 32.0	&70.0	&44.3	&32.1	&72.7	&42.4	&42.0	&32.9&	54.4	&47.0 $\uparrow$\\
\arrayrulecolor{black!100}\bottomrule
\end{tabular}
}
\label{tab: with-or-without-context-lm-eval}
\vspace{-1em}
\end{table}

\subsubsection{Context-Guided Generation}

Can the impact of context be further amplified? Accuracy calculation in Table~\ref{tab: with-or-without-context-lm-eval} relies on log-likelihood that requires probability distributions. Classifier-free guidance modifies the output logits before softmax to steer generation. The logits have been manually shifted, which makes the resulting probability distribution not grounded in the model’s knowledge or data, thus breaking the probabilistic interpretation. Likelihood-base evaluation does not work. We turn to generation-based evaluation instead. Given the limited capacity of our 1.5B base models, we cannot evaluate them meaningfully using free-form generation for the QA tasks. As a result, we assess their ability to generate coherent continuations when provided with a prompt.

\begin{wrapfigure}{rt}{0.4\textwidth}
\vspace{-1em}
    \centering
    \includegraphics[width=\linewidth]{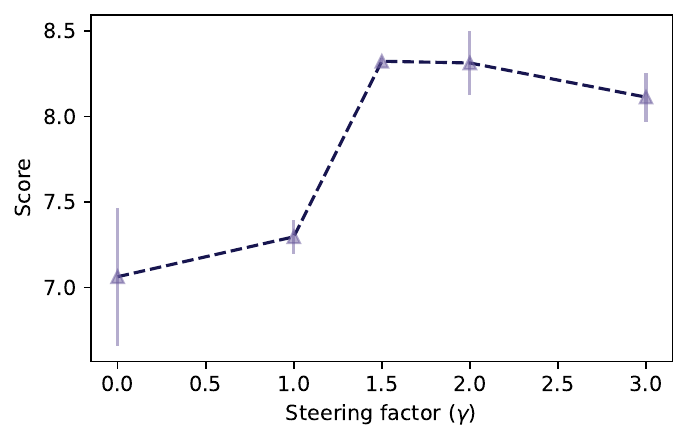}
    \caption{LLM judge scores (with standard deviation) vs. different values of steering factor $\gamma$}
    \label{fig:gamma-sweep}
    \vspace{-1em}
\end{wrapfigure}

We gather prompts in health and history domains, and let our pretrained models to finish the prompts. We let \texttt{gpt-4o} evaluate over 4 dimensions regarding the continuation, that are, coherence, correctness, reasonableness, and relevance. The scores are between 1 and 10, with higher score being more desirable. With context-guided generation, we apply the steering factor $\gamma=1.5$ if not mentioned, from a hyperparameter grid search in Figure~\ref{fig:gamma-sweep}. Over 50 prompts, we have the \texttt{gpt-4o} evaluation presented in Table~\ref{tab:llm-judge-score-prompt-continuation-med-his}, where we compare the three different generation methods using both our context-conditioned pretrained model and standard context-free pretrained model. Given that LLM judges may hallucinate, we run a 3-seed evaluation and report the mean and standard deviation. Compared to the standard pretrained model, we observe that context-conditioned pretrained model is more steerable, as showcased by improved performance with context-guided generation.

\textbf{Which metadata type is efficient in guidance?} Among the three different context types, we notice that even though topic and format domains are not effective in speeding up training, they offer effective steering possibilities. We provide an example of different continuations from the three different sampling methods in Table~\ref{tab:3-different-conts}. Conditioning on \texttt{Topic: Health, Format: Knowledge Article}, while context-free generation offers very generic answers, context-condiitoned generation is able to offer more health-relevant technical answers. Context-guided generation is the most comprehensive and informative among the three, indicating the successful guidance of the context.

\begin{table}[h!]
\caption{Average LLM-as-a-judge evaluation scores (mean $\pm$ std)  on the different prompt completions on 50 prompts on health \& medical and history topics respectively (answer length: 64 tokens). Model type denotes the type of checkpoints used, with "conditioned" being our context-conditioned pretrained checkpoint and "standard" being the standard context-free pretrained checkpoint. For each of the different contexts, we use different context-pretrained models.}
    \centering
    \resizebox{\textwidth}{!}{
    \begin{tabular}{llcccc}
    \toprule
         \textbf{topic} & \textbf{context} & \textbf{model type} &  \textbf{ctx-free} & \textbf{ctx-conditioned} & \textbf{ctx-guided}  \\
         \toprule
          \multirow{6}{*}{health \& medical} &\multirow{2}{*}{\emph{URL: https://medicalxpress.com/}} & {\emph{Conditioned}} &  8.19 $\pm$ 0.14 & 8.14 $\pm$ 0.20 & 8.70 $\pm$ 0.14 \\
         & & \emph{Standard} & 7.98 $\pm$ 0.34 & 7.79 $\pm$ 0.10  & 8.12 $\pm$ 0.26 \\
         \arrayrulecolor{black!30} \cmidrule(lr){2-6}
        & \multirow{2}{*}{\emph{Topic: Health, Format: Knowledge Article}} & \emph{Conditioned} & 8.01 $\pm$ 0.16 & 8.26 $\pm$ 0.14 & \textbf{9.07 $\pm$ 0.05}  \\
        & &  \emph{Standard} & 7.75$\pm$ 0.03 & 7.79 $\pm$ 0.08  & 7.66 $\pm$ 0.24 \\
          \arrayrulecolor{black!30} \cmidrule(lr){2-6}
        & \multirow{2}{*}{\emph{Quality Score: 5}} & \emph{Conditioned} & 8.20 $\pm$ 0.19 & 8.20 $\pm$ 0.20 & 8.69 $\pm$ 0.14 \\
        & &  \emph{Standard} & 8.37 $\pm$ 0.20 & 8.22 $\pm$ 0.23  & 8.09 $\pm$ 0.21 \\
      \arrayrulecolor{black!50} \cmidrule(lr){1-6}
       \multirow{6}{*}{history} & \multirow{2}{*}{\emph{URL: https://britannica.com/}} & \emph{Conditioned} & 5.70 $\pm$ 0.15 & 5.80 $\pm$ 0.16 & 6.73 $\pm$ 0.43 \\
          & & \emph{Standard} & 4.88 $\pm$ 0.15 & 5.60 $\pm$ 0.14 & 6.45 $\pm$ 0.01 \\
          \arrayrulecolor{black!30} \cmidrule(lr){2-6}
        & \multirow{2}{*}{\emph{Topic: History, Format: Knowledge Article}}  & \emph{Conditioned} & 5.35 $\pm$ 0.24 & 5.23 $\pm$ 0.26 & \textbf{7.15 $\pm$ 0.26} \\
        &  &  \emph{Standard} & 5.19 $\pm$ 0.31 & 5.95 $\pm$ 0.11 & 6.04 $\pm$ 0.31 \\
          \arrayrulecolor{black!30} \cmidrule(lr){2-6}
        &  \multirow{2}{*}{\emph{Quality Score: 5}} & \emph{Conditioned} & 5.36 $\pm$ 0.20 & 5.67 $\pm$ 0.16 & 6.45 $\pm$ 0.34\\
       &   &  \emph{Standard} & 5.52 $\pm$ 0.19 & 5.68 $\pm$ 0.19 & 7.04 $\pm$ 0.09 \\
      \arrayrulecolor{black!100} \bottomrule
    \end{tabular}
    }
    \label{tab:llm-judge-score-prompt-continuation-med-his}
    \vspace{-1em}
\end{table}

{\ttfamily
\begin{table}[h!]
\caption{Example of different continuations (answer length: 64 tokens) given a prompt and their corresponding ratings. The Domain Information context provided is \texttt{Topic: Health, Format: Knowledge Article}. Continuations are generated from our DI-conditioned pretrained checkpoints. (Caution of Hallucination!)}
    \centering
    \resizebox{\textwidth}{!}{
   \begin{tabular}{p{0.2\linewidth} p{0.6\linewidth} p{0.2\linewidth}}
\toprule
&  \emph{Given prompt}: Common causes of balance problems in older people include & Rating/Short Rationale   \\
\toprule
standard &  \textcolor{Mulberry}{falls and underlying conditions}. These can range from minor health issues to more serious problems. 
You can get balance problems from several causes, including \textcolor{Mulberry}{poor nutrition, aging, and underlying health conditions like high blood pressure or thyroid problems}. It’s also possible to suffer from balance problems suddenly. &  5.5. Clear but general and slightly repetitive
\\
\midrule
context-conditioned &  \textcolor{Mulberry}{medications, ear problems, and spinal and neck problems}. Vestibular rehabilitation physical therapy, which uses special exercises to enhance brain signals that help maintain balance, can be used to treat some types of balance disorders. In most cases, medical attention is necessary to treat this type of dizziness or imbalance. & 7.0. More technical, introduces helpful treatment concepts
 \\
\midrule
context-guided & \textcolor{Mulberry}{ageing-related changes in the inner ear, neurological problems, and problems with medication}. Other reasons for balance problems include \textcolor{Mulberry}{head injury, Parkinson’s disease, multiple sclerosis, and stroke.
Vitamin B12 deficiency} can cause balance problems. & 9.0. Most comprehensive and informative \\
\bottomrule
    \end{tabular}
    }
    \label{tab:3-different-conts}
\end{table}
}

{\ttfamily
\begin{table}[h!]
\caption{Example of different context-guided continuations given a prompt and a steering factor~($\gamma$). The Domain Information context is \texttt{Topic: Fashion \& Beauty, Format: Personal Blog}. Continuations are generated from our pretrained DI-conditioned checkpoints. (Caution of Hallucination!)}
\centering
\resizebox{\textwidth}{!}{
\begin{tabular}{p{0.2\linewidth} p{0.8\linewidth}}
\toprule
Steering factor ($\gamma$) &  \emph{Given prompt}: Julius Caesar was assassinated because \\
\toprule
-2 & he’d killed too many people and he’d become too dangerous. He was assassinated by senators and officials who were on the wrong side of the Senate. Caesar feared they might take power away from him and start an anarchy. \\
\midrule
0 & he was trying to become king of the Roman Empire. He was forced to flee to Greece to escape the Roman legions. This was a major mistake, because he was already an experienced general and knew that he could not defeat them in the long run. \\
\midrule
1 & his assassins feared he would be a threat to their power. At the time, the average Roman man’s \textcolor{Mulberry}{wardrobe consisted of a tunic and a short-sleeved shirt}. But as soon as Julius Caesar was assassinated, Rome’s \textcolor{Mulberry}{dress code became stricter}. \\
\midrule
2 & of his \textcolor{Mulberry}{taste in makeup}. Back then, \textcolor{Mulberry}{makeup was not only for looks}. It also had medicinal purposes. The Romans \textcolor{Mulberry}{used makeup to soothe and protect the skin from the sun}. It also had medicinal purposes. They believed it \textcolor{Mulberry}{could heal certain skin conditions and infections}. \\
\midrule
4 & he \textcolor{Mulberry}{wore too much jewelry}. Napoleon was so inspired by Alexander the Great \textcolor{Mulberry}{that jewelry became a sign of royalty and he wore very little other than crowns, earrings, necklaces and bracelets}. Kings, Queens, royalty and those with influence \textcolor{Mulberry}{wore so much more than is necessary}. \\
\bottomrule
\end{tabular}
}
\vspace{-1em}
\label{tab:impact-of-gamma}
\end{table}
}

\textbf{Controllable guidance level.} We further demonstrate various steering possibilities enabled by our method. For example, it is possible to guide the generation toward a desired topic by increasing the guidance factor value ($\gamma$). When conditioning on \texttt{Topic: Fashion \& Beauty, Format: Personal Blog} and setting $\gamma > 1$, LLMs sometimes produce humorous or absurd outputs in Figure~\ref{tab:impact-of-gamma}, such as stating that Julius Caesar was assassinated because "of his taste in makeup" or "he wore too much jewelry." While this clearly highlights the model's tendency to hallucinate, it also underscores the strength of the steering mechanism. With appropriate contextual grounding, such guidance could potentially yield coherent and relevant outputs in desired domains.

\textbf{Controllable style and content.} Moreover, when provided with different contextual signals, our models can adapt the continuation in ways that align with the intended framing or source. As shown in Table~\ref{tab:guidance-with-different-URLs}, the open-ended prompt "Paris is known for" leads to varied continuations depending on the guiding URL: \texttt{medicalxpress.com} prompts a medically-oriented response, \texttt{nytimes.com} results in a factual and formal description by listing facts, and \texttt{reddit.com} encourages a casual, entertainment-focused tone. A similar result with Domain Information guidance is presented in Table~\ref{tab:guidance-with-different-topics} and Table~\ref{tab:guidance-with-different-formats} in the Appendix, where we observe effective guidance in both topic and format domains. These results demonstrate that contextual steering can shape both the content and style of the generated text in meaningful ways.

\begin{myblock}
\textbf{Takeaway 3.} With context-conditioned pretraining, a larger improvement is observed in context-guided generation compared to context-conditioned generation.
\end{myblock}

\begin{myblock}
\textbf{Takeaway 4.} Although domain information does not accelerate pretraining, it provides more interpretable and efficient guidance during inference.
\end{myblock}

{\ttfamily
\begin{table}[t!]
\caption{Example of different continuations given a prompt conditioned on different URLs. The continuations are generated in a context-guided fashion using our URL-conditioned pretrained checkpoints. (Caution of Hallucination!)}
    \centering
\resizebox{\textwidth}{!}{
\begin{tabular}{p{0.2\linewidth} p{0.8\linewidth}}
\toprule
& \emph{Given prompt}: Paris is known for \\
\toprule
\url{https://medicalxpress.com/} & 150 years of medical excellence, thanks to the collaboration of doctors, nurses, medical students and researchers. It is also known as the home of major medical discoveries such as the operation of the heart, the discovery of insulin, the discovery of the HIV virus, the discovery of the X-ray, the discovery of the...\\
\midrule
\url{https://nytimes.com/} & 4 things:
1. Big. It’s the capital of France and also the largest city in Europe.
2. Beautiful. It’s a very beautiful city.
3. Rich. It’s one of the richest cities in the world.
4. Old. It’s one of the oldest cities in the world.
 \\
\midrule
\url{https://reddit.com/} & 4 things, its cuisine, its fashion, its nightlife and its cinema. And I don’t mean that only. \smileyface
What if I told you that there is a community of actors who actually act out the scenes of French movies? This community is called REDdit!
 \\
\bottomrule
    \end{tabular}
    }
    \label{tab:guidance-with-different-URLs}
\end{table}
}

\section{Conclusion}
In this work, we present a comprehensive investigation into context-conditioned pretraining for LLMs, focusing on the impact of different metadata types. Our findings reveal that not all metadata types contribute equally to model efficiency or downstream performance. Specifically, URL metadata stands out as the only form of context that significantly accelerates training and improves performance—particularly when longer prompts are used at inference time. In contrast, quality scores and topic/format domain information do not yield measurable benefits during pretraining.

Nevertheless, we demonstrate that these latter forms of metadata have strong value during inference, enabling more controllable and interpretable text generation through classifier-free guidance mechanisms. This dual insight, that certain metadata can optimize training efficiency while others enhance controllability, opens promising directions for more flexible and human-aligned language model applications.

Overall, our results suggest that context-aware pretraining, especially when paired with dynamic guidance strategies at inference, offers a powerful toolset for building more efficient, adaptive, and steerable LLMs. Future work could explore scaling these techniques to larger models and broader metadata sources, as well as integrating contextual signals directly into fine-tuning and instruction tuning pipelines.

\textbf{Acknowledgment.} This work was supported as part of the Swiss AI Initiative by a grant from the Swiss National Supercomputing Centre (CSCS) under project ID a06 on Alps. DF would like to thank Diba Hashemi for helpful discussions.

\newpage
\bibliography{bib}

@misc{allenzhu2024physicslanguagemodels33,
      title={Physics of Language Models: Part 3.3, Knowledge Capacity Scaling Laws}, 
      author={Zeyuan Allen-Zhu and Yuanzhi Li},
      year={2024},
      eprint={2404.05405},
      archivePrefix={arXiv},
      primaryClass={cs.CL},
      url={https://arxiv.org/abs/2404.05405}, 
}

@misc{gao2025metadataconditioningaccelerateslanguage,
      title={Metadata Conditioning Accelerates Language Model Pre-training}, 
      author={Tianyu Gao and Alexander Wettig and Luxi He and Yihe Dong and Sadhika Malladi and Danqi Chen},
      year={2025},
      eprint={2501.01956},
      archivePrefix={arXiv},
      primaryClass={cs.CL},
      url={https://arxiv.org/abs/2501.01956}, 
}

@article{clark2018think,
  title={Think you have solved question answering? try arc, the ai2 reasoning challenge},
  author={Clark, Peter and Cowhey, Isaac and Etzioni, Oren and Khot, Tushar and Sabharwal, Ashish and Schoenick, Carissa and Tafjord, Oyvind},
  journal={arXiv preprint arXiv:1803.05457},
  year={2018}
}

@article{talmor2018commonsenseqa,
  title={Commonsenseqa: A question answering challenge targeting commonsense knowledge},
  author={Talmor, Alon and Herzig, Jonathan and Lourie, Nicholas and Berant, Jonathan},
  journal={arXiv preprint arXiv:1811.00937},
  year={2018}
}

@article{hendrycks2020measuring,
  title={Measuring massive multitask language understanding},
  author={Hendrycks, Dan and Burns, Collin and Basart, Steven and Zou, Andy and Mazeika, Mantas and Song, Dawn and Steinhardt, Jacob},
  journal={arXiv preprint arXiv:2009.03300},
  year={2020}
}

@inproceedings{bisk2020piqa,
  title={Piqa: Reasoning about physical commonsense in natural language},
  author={Bisk, Yonatan and Zellers, Rowan and Gao, Jianfeng and Choi, Yejin and others},
  booktitle={Proceedings of the AAAI conference on artificial intelligence},
  volume={34},
  number={05},
  pages={7432--7439},
  year={2020}
}

@article{sap2019socialiqa,
  title={Socialiqa: Commonsense reasoning about social interactions},
  author={Sap, Maarten and Rashkin, Hannah and Chen, Derek and LeBras, Ronan and Choi, Yejin},
  journal={arXiv preprint arXiv:1904.09728},
  year={2019}
}

@article{zellers2019hellaswag,
  title={Hellaswag: Can a machine really finish your sentence?},
  author={Zellers, Rowan and Holtzman, Ari and Bisk, Yonatan and Farhadi, Ali and Choi, Yejin},
  journal={arXiv preprint arXiv:1905.07830},
  year={2019}
}

@article{paperno2016lambada,
  title={The LAMBADA dataset: Word prediction requiring a broad discourse context},
  author={Paperno, Denis and Kruszewski, Germ{\'a}n and Lazaridou, Angeliki and Pham, Quan Ngoc and Bernardi, Raffaella and Pezzelle, Sandro and Baroni, Marco and Boleda, Gemma and Fern{\'a}ndez, Raquel},
  journal={arXiv preprint arXiv:1606.06031},
  year={2016}
}

@article{sakaguchi2021winogrande,
  title={Winogrande: An adversarial winograd schema challenge at scale},
  author={Sakaguchi, Keisuke and Bras, Ronan Le and Bhagavatula, Chandra and Choi, Yejin},
  journal={Communications of the ACM},
  volume={64},
  number={9},
  pages={99--106},
  year={2021},
  publisher={ACM New York, NY, USA}
}

@misc{higuchi2025doesmetadataconditioningnot,
      title={When Does Metadata Conditioning (NOT) Work for Language Model Pre-Training? A Study with Context-Free Grammars}, 
      author={Rei Higuchi and Ryotaro Kawata and Naoki Nishikawa and Kazusato Oko and Shoichiro Yamaguchi and Sosuke Kobayashi and Seiya Tokui and Kohei Hayashi and Daisuke Okanohara and Taiji Suzuki},
      year={2025},
      eprint={2504.17562},
      archivePrefix={arXiv},
      primaryClass={cs.CL},
      url={https://arxiv.org/abs/2504.17562}, 
}

@misc{eval-harness,
  author       = {Gao, Leo and Tow, Jonathan and Abbasi, Baber and Biderman, Stella and Black, Sid and DiPofi, Anthony and Foster, Charles and Golding, Laurence and Hsu, Jeffrey and Le Noac'h, Alain and Li, Haonan and McDonell, Kyle and Muennighoff, Niklas and Ociepa, Chris and Phang, Jason and Reynolds, Laria and Schoelkopf, Hailey and Skowron, Aviya and Sutawika, Lintang and Tang, Eric and Thite, Anish and Wang, Ben and Wang, Kevin and Zou, Andy},
  title        = {A framework for few-shot language model evaluation},
  month        = 07,
  year         = 2024,
  publisher    = {Zenodo},
  version      = {v0.4.3},
  doi          = {10.5281/zenodo.12608602},
  url          = {https://zenodo.org/records/12608602}
}

@misc{ho2022classifierfreediffusionguidance,
      title={Classifier-Free Diffusion Guidance}, 
      author={Jonathan Ho and Tim Salimans},
      year={2022},
      eprint={2207.12598},
      archivePrefix={arXiv},
      primaryClass={cs.LG},
      url={https://arxiv.org/abs/2207.12598}, 
}

@misc{wettig2025organizewebconstructingdomains,
      title={Organize the Web: Constructing Domains Enhances Pre-Training Data Curation}, 
      author={Alexander Wettig and Kyle Lo and Sewon Min and Hannaneh Hajishirzi and Danqi Chen and Luca Soldaini},
      year={2025},
      eprint={2502.10341},
      archivePrefix={arXiv},
      primaryClass={cs.CL},
      url={https://arxiv.org/abs/2502.10341}, 
}

@misc{grattafiori2024llama3herdmodels,
      title={The Llama 3 Herd of Models}, 
      author={Aaron Grattafiori and Abhimanyu Dubey and Abhinav Jauhri and Abhinav Pandey and Abhishek Kadian and Ahmad Al-Dahle and Aiesha Letman and Akhil Mathur and Alan Schelten and Alex Vaughan and Amy Yang and Angela Fan and Anirudh Goyal and Anthony Hartshorn and Aobo Yang and Archi Mitra and Archie Sravankumar and Artem Korenev and Arthur Hinsvark and Arun Rao and Aston Zhang and Aurelien Rodriguez and Austen Gregerson and Ava Spataru and Baptiste Roziere and Bethany Biron and Binh Tang and Bobbie Chern and Charlotte Caucheteux and Chaya Nayak and Chloe Bi and Chris Marra and Chris McConnell and Christian Keller and Christophe Touret and Chunyang Wu and Corinne Wong and Cristian Canton Ferrer and Cyrus Nikolaidis and Damien Allonsius and Daniel Song and Danielle Pintz and Danny Livshits and Danny Wyatt and David Esiobu and Dhruv Choudhary and Dhruv Mahajan and Diego Garcia-Olano and Diego Perino and Dieuwke Hupkes and Egor Lakomkin and Ehab AlBadawy and Elina Lobanova and Emily Dinan and Eric Michael Smith and Filip Radenovic and Francisco Guzmán and Frank Zhang and Gabriel Synnaeve and Gabrielle Lee and Georgia Lewis Anderson and Govind Thattai and Graeme Nail and Gregoire Mialon and Guan Pang and Guillem Cucurell and Hailey Nguyen and Hannah Korevaar and Hu Xu and Hugo Touvron and Iliyan Zarov and Imanol Arrieta Ibarra and Isabel Kloumann and Ishan Misra and Ivan Evtimov and Jack Zhang and Jade Copet and Jaewon Lee and Jan Geffert and Jana Vranes and Jason Park and Jay Mahadeokar and Jeet Shah and Jelmer van der Linde and Jennifer Billock and Jenny Hong and Jenya Lee and Jeremy Fu and Jianfeng Chi and Jianyu Huang and Jiawen Liu and Jie Wang and Jiecao Yu and Joanna Bitton and Joe Spisak and Jongsoo Park and Joseph Rocca and Joshua Johnstun and Joshua Saxe and Junteng Jia and Kalyan Vasuden Alwala and Karthik Prasad and Kartikeya Upasani and Kate Plawiak and Ke Li and Kenneth Heafield and Kevin Stone and Khalid El-Arini and Krithika Iyer and Kshitiz Malik and Kuenley Chiu and Kunal Bhalla and Kushal Lakhotia and Lauren Rantala-Yeary and Laurens van der Maaten and Lawrence Chen and Liang Tan and Liz Jenkins and Louis Martin and Lovish Madaan and Lubo Malo and Lukas Blecher and Lukas Landzaat and Luke de Oliveira and Madeline Muzzi and Mahesh Pasupuleti and Mannat Singh and Manohar Paluri and Marcin Kardas and Maria Tsimpoukelli and Mathew Oldham and Mathieu Rita and Maya Pavlova and Melanie Kambadur and Mike Lewis and Min Si and Mitesh Kumar Singh and Mona Hassan and Naman Goyal and Narjes Torabi and Nikolay Bashlykov and Nikolay Bogoychev and Niladri Chatterji and Ning Zhang and Olivier Duchenne and Onur Çelebi and Patrick Alrassy and Pengchuan Zhang and Pengwei Li and Petar Vasic and Peter Weng and Prajjwal Bhargava and Pratik Dubal and Praveen Krishnan and Punit Singh Koura and Puxin Xu and Qing He and Qingxiao Dong and Ragavan Srinivasan and Raj Ganapathy and Ramon Calderer and Ricardo Silveira Cabral and Robert Stojnic and Roberta Raileanu and Rohan Maheswari and Rohit Girdhar and Rohit Patel and Romain Sauvestre and Ronnie Polidoro and Roshan Sumbaly and Ross Taylor and Ruan Silva and Rui Hou and Rui Wang and Saghar Hosseini and Sahana Chennabasappa and Sanjay Singh and Sean Bell and Seohyun Sonia Kim and Sergey Edunov and Shaoliang Nie and Sharan Narang and Sharath Raparthy and Sheng Shen and Shengye Wan and Shruti Bhosale and Shun Zhang and Simon Vandenhende and Soumya Batra and Spencer Whitman and Sten Sootla and Stephane Collot and Suchin Gururangan and Sydney Borodinsky and Tamar Herman and Tara Fowler and Tarek Sheasha and Thomas Georgiou and Thomas Scialom and Tobias Speckbacher and Todor Mihaylov and Tong Xiao and Ujjwal Karn and Vedanuj Goswami and Vibhor Gupta and Vignesh Ramanathan and Viktor Kerkez and Vincent Gonguet and Virginie Do and Vish Vogeti and Vítor Albiero and Vladan Petrovic and Weiwei Chu and Wenhan Xiong and Wenyin Fu and Whitney Meers and Xavier Martinet and Xiaodong Wang and Xiaofang Wang and Xiaoqing Ellen Tan and Xide Xia and Xinfeng Xie and Xuchao Jia and Xuewei Wang and Yaelle Goldschlag and Yashesh Gaur and Yasmine Babaei and Yi Wen and Yiwen Song and Yuchen Zhang and Yue Li and Yuning Mao and Zacharie Delpierre Coudert and Zheng Yan and Zhengxing Chen and Zoe Papakipos and Aaditya Singh and Aayushi Srivastava and Abha Jain and Adam Kelsey and Adam Shajnfeld and Adithya Gangidi and Adolfo Victoria and Ahuva Goldstand and Ajay Menon and Ajay Sharma and Alex Boesenberg and Alexei Baevski and Allie Feinstein and Amanda Kallet and Amit Sangani and Amos Teo and Anam Yunus and Andrei Lupu and Andres Alvarado and Andrew Caples and Andrew Gu and Andrew Ho and Andrew Poulton and Andrew Ryan and Ankit Ramchandani and Annie Dong and Annie Franco and Anuj Goyal and Aparajita Saraf and Arkabandhu Chowdhury and Ashley Gabriel and Ashwin Bharambe and Assaf Eisenman and Azadeh Yazdan and Beau James and Ben Maurer and Benjamin Leonhardi and Bernie Huang and Beth Loyd and Beto De Paola and Bhargavi Paranjape and Bing Liu and Bo Wu and Boyu Ni and Braden Hancock and Bram Wasti and Brandon Spence and Brani Stojkovic and Brian Gamido and Britt Montalvo and Carl Parker and Carly Burton and Catalina Mejia and Ce Liu and Changhan Wang and Changkyu Kim and Chao Zhou and Chester Hu and Ching-Hsiang Chu and Chris Cai and Chris Tindal and Christoph Feichtenhofer and Cynthia Gao and Damon Civin and Dana Beaty and Daniel Kreymer and Daniel Li and David Adkins and David Xu and Davide Testuggine and Delia David and Devi Parikh and Diana Liskovich and Didem Foss and Dingkang Wang and Duc Le and Dustin Holland and Edward Dowling and Eissa Jamil and Elaine Montgomery and Eleonora Presani and Emily Hahn and Emily Wood and Eric-Tuan Le and Erik Brinkman and Esteban Arcaute and Evan Dunbar and Evan Smothers and Fei Sun and Felix Kreuk and Feng Tian and Filippos Kokkinos and Firat Ozgenel and Francesco Caggioni and Frank Kanayet and Frank Seide and Gabriela Medina Florez and Gabriella Schwarz and Gada Badeer and Georgia Swee and Gil Halpern and Grant Herman and Grigory Sizov and Guangyi and Zhang and Guna Lakshminarayanan and Hakan Inan and Hamid Shojanazeri and Han Zou and Hannah Wang and Hanwen Zha and Haroun Habeeb and Harrison Rudolph and Helen Suk and Henry Aspegren and Hunter Goldman and Hongyuan Zhan and Ibrahim Damlaj and Igor Molybog and Igor Tufanov and Ilias Leontiadis and Irina-Elena Veliche and Itai Gat and Jake Weissman and James Geboski and James Kohli and Janice Lam and Japhet Asher and Jean-Baptiste Gaya and Jeff Marcus and Jeff Tang and Jennifer Chan and Jenny Zhen and Jeremy Reizenstein and Jeremy Teboul and Jessica Zhong and Jian Jin and Jingyi Yang and Joe Cummings and Jon Carvill and Jon Shepard and Jonathan McPhie and Jonathan Torres and Josh Ginsburg and Junjie Wang and Kai Wu and Kam Hou U and Karan Saxena and Kartikay Khandelwal and Katayoun Zand and Kathy Matosich and Kaushik Veeraraghavan and Kelly Michelena and Keqian Li and Kiran Jagadeesh and Kun Huang and Kunal Chawla and Kyle Huang and Lailin Chen and Lakshya Garg and Lavender A and Leandro Silva and Lee Bell and Lei Zhang and Liangpeng Guo and Licheng Yu and Liron Moshkovich and Luca Wehrstedt and Madian Khabsa and Manav Avalani and Manish Bhatt and Martynas Mankus and Matan Hasson and Matthew Lennie and Matthias Reso and Maxim Groshev and Maxim Naumov and Maya Lathi and Meghan Keneally and Miao Liu and Michael L. Seltzer and Michal Valko and Michelle Restrepo and Mihir Patel and Mik Vyatskov and Mikayel Samvelyan and Mike Clark and Mike Macey and Mike Wang and Miquel Jubert Hermoso and Mo Metanat and Mohammad Rastegari and Munish Bansal and Nandhini Santhanam and Natascha Parks and Natasha White and Navyata Bawa and Nayan Singhal and Nick Egebo and Nicolas Usunier and Nikhil Mehta and Nikolay Pavlovich Laptev and Ning Dong and Norman Cheng and Oleg Chernoguz and Olivia Hart and Omkar Salpekar and Ozlem Kalinli and Parkin Kent and Parth Parekh and Paul Saab and Pavan Balaji and Pedro Rittner and Philip Bontrager and Pierre Roux and Piotr Dollar and Polina Zvyagina and Prashant Ratanchandani and Pritish Yuvraj and Qian Liang and Rachad Alao and Rachel Rodriguez and Rafi Ayub and Raghotham Murthy and Raghu Nayani and Rahul Mitra and Rangaprabhu Parthasarathy and Raymond Li and Rebekkah Hogan and Robin Battey and Rocky Wang and Russ Howes and Ruty Rinott and Sachin Mehta and Sachin Siby and Sai Jayesh Bondu and Samyak Datta and Sara Chugh and Sara Hunt and Sargun Dhillon and Sasha Sidorov and Satadru Pan and Saurabh Mahajan and Saurabh Verma and Seiji Yamamoto and Sharadh Ramaswamy and Shaun Lindsay and Shaun Lindsay and Sheng Feng and Shenghao Lin and Shengxin Cindy Zha and Shishir Patil and Shiva Shankar and Shuqiang Zhang and Shuqiang Zhang and Sinong Wang and Sneha Agarwal and Soji Sajuyigbe and Soumith Chintala and Stephanie Max and Stephen Chen and Steve Kehoe and Steve Satterfield and Sudarshan Govindaprasad and Sumit Gupta and Summer Deng and Sungmin Cho and Sunny Virk and Suraj Subramanian and Sy Choudhury and Sydney Goldman and Tal Remez and Tamar Glaser and Tamara Best and Thilo Koehler and Thomas Robinson and Tianhe Li and Tianjun Zhang and Tim Matthews and Timothy Chou and Tzook Shaked and Varun Vontimitta and Victoria Ajayi and Victoria Montanez and Vijai Mohan and Vinay Satish Kumar and Vishal Mangla and Vlad Ionescu and Vlad Poenaru and Vlad Tiberiu Mihailescu and Vladimir Ivanov and Wei Li and Wenchen Wang and Wenwen Jiang and Wes Bouaziz and Will Constable and Xiaocheng Tang and Xiaojian Wu and Xiaolan Wang and Xilun Wu and Xinbo Gao and Yaniv Kleinman and Yanjun Chen and Ye Hu and Ye Jia and Ye Qi and Yenda Li and Yilin Zhang and Ying Zhang and Yossi Adi and Youngjin Nam and Yu and Wang and Yu Zhao and Yuchen Hao and Yundi Qian and Yunlu Li and Yuzi He and Zach Rait and Zachary DeVito and Zef Rosnbrick and Zhaoduo Wen and Zhenyu Yang and Zhiwei Zhao and Zhiyu Ma},
      year={2024},
      eprint={2407.21783},
      archivePrefix={arXiv},
      primaryClass={cs.AI},
      url={https://arxiv.org/abs/2407.21783}, 
}

@misc{loshchilov2019decoupledweightdecayregularization,
      title={Decoupled Weight Decay Regularization}, 
      author={Ilya Loshchilov and Frank Hutter},
      year={2019},
      eprint={1711.05101},
      archivePrefix={arXiv},
      primaryClass={cs.LG},
      url={https://arxiv.org/abs/1711.05101}, 
}

@misc{narayanan2021efficientlargescalelanguagemodel,
      title={Efficient Large-Scale Language Model Training on GPU Clusters Using Megatron-LM}, 
      author={Deepak Narayanan and Mohammad Shoeybi and Jared Casper and Patrick LeGresley and Mostofa Patwary and Vijay Anand Korthikanti and Dmitri Vainbrand and Prethvi Kashinkunti and Julie Bernauer and Bryan Catanzaro and Amar Phanishayee and Matei Zaharia},
      year={2021},
      eprint={2104.04473},
      archivePrefix={arXiv},
      primaryClass={cs.CL},
      url={https://arxiv.org/abs/2104.04473}, 
}

@misc{lozhkov2024fineweb-edu,
    author       = { Lozhkov, Anton and Ben Allal, Loubna and von Werra, Leandro and Wolf, Thomas },  
    title        = { FineWeb-Edu: the Finest Collection of Educational Content }, 
    year         = 2024,  
    url          = { https://huggingface.co/datasets/HuggingFaceFW/fineweb-edu },  
    doi          = { 10.57967/hf/2497 },
    publisher    = { Hugging Face }
}

@misc{he2025contextsteeringcontrollablepersonalization,
      title={Context Steering: Controllable Personalization at Inference Time}, 
      author={Jerry Zhi-Yang He and Sashrika Pandey and Mariah L. Schrum and Anca Dragan},
      year={2025},
      eprint={2405.01768},
      archivePrefix={arXiv},
      primaryClass={cs.CL},
      url={https://arxiv.org/abs/2405.01768}, 
}

@misc{sanchez2023staytopicclassifierfreeguidance,
      title={Stay on topic with Classifier-Free Guidance}, 
      author={Guillaume Sanchez and Honglu Fan and Alexander Spangher and Elad Levi and Pawan Sasanka Ammanamanchi and Stella Biderman},
      year={2023},
      eprint={2306.17806},
      archivePrefix={arXiv},
      primaryClass={cs.CL},
      url={https://arxiv.org/abs/2306.17806}, 
}

@inproceedings{gu-etal-2025-olmes,
    title = "{OLMES}: A Standard for Language Model Evaluations",
    author = "Gu, Yuling  and
      Tafjord, Oyvind  and
      Kuehl, Bailey  and
      Haddad, Dany  and
      Dodge, Jesse  and
      Hajishirzi, Hannaneh",
    editor = "Chiruzzo, Luis  and
      Ritter, Alan  and
      Wang, Lu",
    booktitle = "Findings of the Association for Computational Linguistics: NAACL 2025",
    month = apr,
    year = "2025",
    address = "Albuquerque, New Mexico",
    publisher = "Association for Computational Linguistics",
    url = "https://aclanthology.org/2025.findings-naacl.282/",
    pages = "5005--5033",
    ISBN = "979-8-89176-195-7"
}

@misc{zhu2025powercontextenhancedlearningllms,
      title={On the Power of Context-Enhanced Learning in LLMs}, 
      author={Xingyu Zhu and Abhishek Panigrahi and Sanjeev Arora},
      year={2025},
      eprint={2503.01821},
      archivePrefix={arXiv},
      primaryClass={cs.LG},
      url={https://arxiv.org/abs/2503.01821}, 
}

@inproceedings{li-etal-2023-contrastive,
    title = "Contrastive Decoding: Open-ended Text Generation as Optimization",
    author = "Li, Xiang Lisa  and
      Holtzman, Ari  and
      Fried, Daniel  and
      Liang, Percy  and
      Eisner, Jason  and
      Hashimoto, Tatsunori  and
      Zettlemoyer, Luke  and
      Lewis, Mike",
    editor = "Rogers, Anna  and
      Boyd-Graber, Jordan  and
      Okazaki, Naoaki",
    booktitle = "Proceedings of the 61st Annual Meeting of the Association for Computational Linguistics (Volume 1: Long Papers)",
    month = jul,
    year = "2023",
    address = "Toronto, Canada",
    publisher = "Association for Computational Linguistics",
    url = "https://aclanthology.org/2023.acl-long.687/",
    doi = "10.18653/v1/2023.acl-long.687",
    pages = "12286--12312"
}

@inproceedings{liu-etal-2021-dexperts,
    title = "{DE}xperts: Decoding-Time Controlled Text Generation with Experts and Anti-Experts",
    author = "Liu, Alisa  and
      Sap, Maarten  and
      Lu, Ximing  and
      Swayamdipta, Swabha  and
      Bhagavatula, Chandra  and
      Smith, Noah A.  and
      Choi, Yejin",
    editor = "Zong, Chengqing  and
      Xia, Fei  and
      Li, Wenjie  and
      Navigli, Roberto",
    booktitle = "Proceedings of the 59th Annual Meeting of the Association for Computational Linguistics and the 11th International Joint Conference on Natural Language Processing (Volume 1: Long Papers)",
    month = aug,
    year = "2021",
    address = "Online",
    publisher = "Association for Computational Linguistics",
    url = "https://aclanthology.org/2021.acl-long.522/",
    doi = "10.18653/v1/2021.acl-long.522",
    pages = "6691--6706"
}

@inproceedings{
du2024understanding,
title={Understanding Emergent Abilities of Language Models from the Loss Perspective},
author={Zhengxiao Du and Aohan Zeng and Yuxiao Dong and Jie Tang},
booktitle={The Thirty-eighth Annual Conference on Neural Information Processing Systems},
year={2024},
url={https://openreview.net/forum?id=35DAviqMFo}
}
\bibliographystyle{unsrtnat}

\newpage
\section*{NeurIPS Paper Checklist}

\begin{enumerate}

\item {\bf Claims}
    \item[] Question: Do the main claims made in the abstract and introduction accurately reflect the paper's contributions and scope?
    \item[] Answer: \answerYes{} %
    \item[] Justification: Our main text corresponds to what we claim in the abstract and introduction, i.e., what benefits does context-aware pretraining bring in pretraining and inference time, respectively.
    \item[] Guidelines:
    \begin{itemize}
        \item The answer NA means that the abstract and introduction do not include the claims made in the paper.
        \item The abstract and/or introduction should clearly state the claims made, including the contributions made in the paper and important assumptions and limitations. A No or NA answer to this question will not be perceived well by the reviewers. 
        \item The claims made should match theoretical and experimental results, and reflect how much the results can be expected to generalize to other settings. 
        \item It is fine to include aspirational goals as motivation as long as it is clear that these goals are not attained by the paper. 
    \end{itemize}

\item {\bf Limitations}
    \item[] Question: Does the paper discuss the limitations of the work performed by the authors?
    \item[] Answer: \answerYes{} %
    \item[] Justification: See our discussions in the conclusion section as well as in Appendix~\ref{app: limitations}.
    \item[] Guidelines:
    \begin{itemize}
        \item The answer NA means that the paper has no limitation while the answer No means that the paper has limitations, but those are not discussed in the paper. 
        \item The authors are encouraged to create a separate "Limitations" section in their paper.
        \item The paper should point out any strong assumptions and how robust the results are to violations of these assumptions (e.g., independence assumptions, noiseless settings, model well-specification, asymptotic approximations only holding locally). The authors should reflect on how these assumptions might be violated in practice and what the implications would be.
        \item The authors should reflect on the scope of the claims made, e.g., if the approach was only tested on a few datasets or with a few runs. In general, empirical results often depend on implicit assumptions, which should be articulated.
        \item The authors should reflect on the factors that influence the performance of the approach. For example, a facial recognition algorithm may perform poorly when image resolution is low or images are taken in low lighting. Or a speech-to-text system might not be used reliably to provide closed captions for online lectures because it fails to handle technical jargon.
        \item The authors should discuss the computational efficiency of the proposed algorithms and how they scale with dataset size.
        \item If applicable, the authors should discuss possible limitations of their approach to address problems of privacy and fairness.
        \item While the authors might fear that complete honesty about limitations might be used by reviewers as grounds for rejection, a worse outcome might be that reviewers discover limitations that aren't acknowledged in the paper. The authors should use their best judgment and recognize that individual actions in favor of transparency play an important role in developing norms that preserve the integrity of the community. Reviewers will be specifically instructed to not penalize honesty concerning limitations.
    \end{itemize}

\item {\bf Theory assumptions and proofs}
    \item[] Question: For each theoretical result, does the paper provide the full set of assumptions and a complete (and correct) proof?
    \item[] Answer: \answerNA{} %
    \item[] Justification: The paper does not include theoretical results
    \item[] Guidelines:
    \begin{itemize}
        \item The answer NA means that the paper does not include theoretical results. 
        \item All the theorems, formulas, and proofs in the paper should be numbered and cross-referenced.
        \item All assumptions should be clearly stated or referenced in the statement of any theorems.
        \item The proofs can either appear in the main paper or the supplemental material, but if they appear in the supplemental material, the authors are encouraged to provide a short proof sketch to provide intuition. 
        \item Inversely, any informal proof provided in the core of the paper should be complemented by formal proofs provided in appendix or supplemental material.
        \item Theorems and Lemmas that the proof relies upon should be properly referenced. 
    \end{itemize}

    \item {\bf Experimental result reproducibility}
    \item[] Question: Does the paper fully disclose all the information needed to reproduce the main experimental results of the paper to the extent that it affects the main claims and/or conclusions of the paper (regardless of whether the code and data are provided or not)?
    \item[] Answer: \answerYes{} %
    \item[] Justification: We detailed our tokenization process and metadata used. Moreover, the training details are also specified in Section~\ref{sec: experiments}. 
    \item[] Guidelines:
    \begin{itemize}
        \item The answer NA means that the paper does not include experiments.
        \item If the paper includes experiments, a No answer to this question will not be perceived well by the reviewers: Making the paper reproducible is important, regardless of whether the code and data are provided or not.
        \item If the contribution is a dataset and/or model, the authors should describe the steps taken to make their results reproducible or verifiable. 
        \item Depending on the contribution, reproducibility can be accomplished in various ways. For example, if the contribution is a novel architecture, describing the architecture fully might suffice, or if the contribution is a specific model and empirical evaluation, it may be necessary to either make it possible for others to replicate the model with the same dataset, or provide access to the model. In general. releasing code and data is often one good way to accomplish this, but reproducibility can also be provided via detailed instructions for how to replicate the results, access to a hosted model (e.g., in the case of a large language model), releasing of a model checkpoint, or other means that are appropriate to the research performed.
        \item While NeurIPS does not require releasing code, the conference does require all submissions to provide some reasonable avenue for reproducibility, which may depend on the nature of the contribution. For example
        \begin{enumerate}
            \item If the contribution is primarily a new algorithm, the paper should make it clear how to reproduce that algorithm.
            \item If the contribution is primarily a new model architecture, the paper should describe the architecture clearly and fully.
            \item If the contribution is a new model (e.g., a large language model), then there should either be a way to access this model for reproducing the results or a way to reproduce the model (e.g., with an open-source dataset or instructions for how to construct the dataset).
            \item We recognize that reproducibility may be tricky in some cases, in which case authors are welcome to describe the particular way they provide for reproducibility. In the case of closed-source models, it may be that access to the model is limited in some way (e.g., to registered users), but it should be possible for other researchers to have some path to reproducing or verifying the results.
        \end{enumerate}
    \end{itemize}

\item {\bf Open access to data and code}
    \item[] Question: Does the paper provide open access to the data and code, with sufficient instructions to faithfully reproduce the main experimental results, as described in supplemental material?
    \item[] Answer: \answerNo{} %
    \item[] Justification: We will make our training codes public when the paper is made public. The dataset we used is public.
    \item[] Guidelines:
    \begin{itemize}
        \item The answer NA means that paper does not include experiments requiring code.
        \item Please see the NeurIPS code and data submission guidelines (\url{https://nips.cc/public/guides/CodeSubmissionPolicy}) for more details.
        \item While we encourage the release of code and data, we understand that this might not be possible, so “No” is an acceptable answer. Papers cannot be rejected simply for not including code, unless this is central to the contribution (e.g., for a new open-source benchmark).
        \item The instructions should contain the exact command and environment needed to run to reproduce the results. See the NeurIPS code and data submission guidelines (\url{https://nips.cc/public/guides/CodeSubmissionPolicy}) for more details.
        \item The authors should provide instructions on data access and preparation, including how to access the raw data, preprocessed data, intermediate data, and generated data, etc.
        \item The authors should provide scripts to reproduce all experimental results for the new proposed method and baselines. If only a subset of experiments are reproducible, they should state which ones are omitted from the script and why.
        \item At submission time, to preserve anonymity, the authors should release anonymized versions (if applicable).
        \item Providing as much information as possible in supplemental material (appended to the paper) is recommended, but including URLs to data and code is permitted.
    \end{itemize}

\item {\bf Experimental setting/details}
    \item[] Question: Does the paper specify all the training and test details (e.g., data splits, hyperparameters, how they were chosen, type of optimizer, etc.) necessary to understand the results?
    \item[] Answer: \answerYes{} %
    \item[] Justification: We specified our dataset, hyperparameter choices, and training details in Section~\ref{sec: experiments}.
    \item[] Guidelines:
    \begin{itemize}
        \item The answer NA means that the paper does not include experiments.
        \item The experimental setting should be presented in the core of the paper to a level of detail that is necessary to appreciate the results and make sense of them.
        \item The full details can be provided either with the code, in appendix, or as supplemental material.
    \end{itemize}

\item {\bf Experiment statistical significance}
    \item[] Question: Does the paper report error bars suitably and correctly defined or other appropriate information about the statistical significance of the experiments?
    \item[] Answer: \answerYes{} %
    \item[] Justification: We reported 3-seed run results with both mean and standard deviation in Section~\ref{sec: context-aware generation}.
    \item[] Guidelines:
    \begin{itemize}
        \item The answer NA means that the paper does not include experiments.
        \item The authors should answer "Yes" if the results are accompanied by error bars, confidence intervals, or statistical significance tests, at least for the experiments that support the main claims of the paper.
        \item The factors of variability that the error bars are capturing should be clearly stated (for example, train/test split, initialization, random drawing of some parameter, or overall run with given experimental conditions).
        \item The method for calculating the error bars should be explained (closed form formula, call to a library function, bootstrap, etc.)
        \item The assumptions made should be given (e.g., Normally distributed errors).
        \item It should be clear whether the error bar is the standard deviation or the standard error of the mean.
        \item It is OK to report 1-sigma error bars, but one should state it. The authors should preferably report a 2-sigma error bar than state that they have a 96\% CI, if the hypothesis of Normality of errors is not verified.
        \item For asymmetric distributions, the authors should be careful not to show in tables or figures symmetric error bars that would yield results that are out of range (e.g. negative error rates).
        \item If error bars are reported in tables or plots, The authors should explain in the text how they were calculated and reference the corresponding figures or tables in the text.
    \end{itemize}

\item {\bf Experiments compute resources}
    \item[] Question: For each experiment, does the paper provide sufficient information on the computer resources (type of compute workers, memory, time of execution) needed to reproduce the experiments?
    \item[] Answer: \answerYes{} %
    \item[] Justification: We specified in the Appendix.
    \item[] Guidelines:
    \begin{itemize}
        \item The answer NA means that the paper does not include experiments.
        \item The paper should indicate the type of compute workers CPU or GPU, internal cluster, or cloud provider, including relevant memory and storage.
        \item The paper should provide the amount of compute required for each of the individual experimental runs as well as estimate the total compute. 
        \item The paper should disclose whether the full research project required more compute than the experiments reported in the paper (e.g., preliminary or failed experiments that didn't make it into the paper). 
    \end{itemize}
    
\item {\bf Code of ethics}
    \item[] Question: Does the research conducted in the paper conform, in every respect, with the NeurIPS Code of Ethics \url{https://neurips.cc/public/EthicsGuidelines}?
    \item[] Answer: \answerYes{} %
    \item[] Justification: We adhere to the Code of Ethics.
    \item[] Guidelines:
    \begin{itemize}
        \item The answer NA means that the authors have not reviewed the NeurIPS Code of Ethics.
        \item If the authors answer No, they should explain the special circumstances that require a deviation from the Code of Ethics.
        \item The authors should make sure to preserve anonymity (e.g., if there is a special consideration due to laws or regulations in their jurisdiction).
    \end{itemize}

\item {\bf Broader impacts}
    \item[] Question: Does the paper discuss both potential positive societal impacts and negative societal impacts of the work performed?
    \item[] Answer: \answerYes{} %
    \item[] Justification: See Appendix~\ref{app: broader-impacts}.
    \item[] Guidelines:
    \begin{itemize}
        \item The answer NA means that there is no societal impact of the work performed.
        \item If the authors answer NA or No, they should explain why their work has no societal impact or why the paper does not address societal impact.
        \item Examples of negative societal impacts include potential malicious or unintended uses (e.g., disinformation, generating fake profiles, surveillance), fairness considerations (e.g., deployment of technologies that could make decisions that unfairly impact specific groups), privacy considerations, and security considerations.
        \item The conference expects that many papers will be foundational research and not tied to particular applications, let alone deployments. However, if there is a direct path to any negative applications, the authors should point it out. For example, it is legitimate to point out that an improvement in the quality of generative models could be used to generate deepfakes for disinformation. On the other hand, it is not needed to point out that a generic algorithm for optimizing neural networks could enable people to train models that generate Deepfakes faster.
        \item The authors should consider possible harms that could arise when the technology is being used as intended and functioning correctly, harms that could arise when the technology is being used as intended but gives incorrect results, and harms following from (intentional or unintentional) misuse of the technology.
        \item If there are negative societal impacts, the authors could also discuss possible mitigation strategies (e.g., gated release of models, providing defenses in addition to attacks, mechanisms for monitoring misuse, mechanisms to monitor how a system learns from feedback over time, improving the efficiency and accessibility of ML).
    \end{itemize}
    
\item {\bf Safeguards}
    \item[] Question: Does the paper describe safeguards that have been put in place for responsible release of data or models that have a high risk for misuse (e.g., pretrained language models, image generators, or scraped datasets)?
    \item[] Answer: \answerNA{} %
    \item[] Justification: We are not releasing new models, but rather providing insights to assist model pretraining.
    \item[] Guidelines:
    \begin{itemize}
        \item The answer NA means that the paper poses no such risks.
        \item Released models that have a high risk for misuse or dual-use should be released with necessary safeguards to allow for controlled use of the model, for example by requiring that users adhere to usage guidelines or restrictions to access the model or implementing safety filters. 
        \item Datasets that have been scraped from the Internet could pose safety risks. The authors should describe how they avoided releasing unsafe images.
        \item We recognize that providing effective safeguards is challenging, and many papers do not require this, but we encourage authors to take this into account and make a best faith effort.
    \end{itemize}

\item {\bf Licenses for existing assets}
    \item[] Question: Are the creators or original owners of assets (e.g., code, data, models), used in the paper, properly credited and are the license and terms of use explicitly mentioned and properly respected?
    \item[] Answer: \answerYes{} %
    \item[] Justification: We cited the dataset, model training framework, codebase and tokenizer. 
    \item[] Guidelines:
    \begin{itemize}
        \item The answer NA means that the paper does not use existing assets.
        \item The authors should cite the original paper that produced the code package or dataset.
        \item The authors should state which version of the asset is used and, if possible, include a URL.
        \item The name of the license (e.g., CC-BY 4.0) should be included for each asset.
        \item For scraped data from a particular source (e.g., website), the copyright and terms of service of that source should be provided.
        \item If assets are released, the license, copyright information, and terms of use in the package should be provided. For popular datasets, \url{paperswithcode.com/datasets} has curated licenses for some datasets. Their licensing guide can help determine the license of a dataset.
        \item For existing datasets that are re-packaged, both the original license and the license of the derived asset (if it has changed) should be provided.
        \item If this information is not available online, the authors are encouraged to reach out to the asset's creators.
    \end{itemize}

\item {\bf New assets}
    \item[] Question: Are new assets introduced in the paper well documented and is the documentation provided alongside the assets?
    \item[] Answer: \answerNA{}%
    \item[] Justification: The paper does not release new assets.
    \item[] Guidelines:
    \begin{itemize}
        \item The answer NA means that the paper does not release new assets.
        \item Researchers should communicate the details of the dataset/code/model as part of their submissions via structured templates. This includes details about training, license, limitations, etc. 
        \item The paper should discuss whether and how consent was obtained from people whose asset is used.
        \item At submission time, remember to anonymize your assets (if applicable). You can either create an anonymized URL or include an anonymized zip file.
    \end{itemize}

\item {\bf Crowdsourcing and research with human subjects}
    \item[] Question: For crowdsourcing experiments and research with human subjects, does the paper include the full text of instructions given to participants and screenshots, if applicable, as well as details about compensation (if any)? 
    \item[] Answer: \answerNA{}%
    \item[] Justification: 
    The paper does not involve crowdsourcing nor research with human subjects.
    \item[] Guidelines:
    \begin{itemize}
        \item The answer NA means that the paper does not involve crowdsourcing nor research with human subjects.
        \item Including this information in the supplemental material is fine, but if the main contribution of the paper involves human subjects, then as much detail as possible should be included in the main paper. 
        \item According to the NeurIPS Code of Ethics, workers involved in data collection, curation, or other labor should be paid at least the minimum wage in the country of the data collector. 
    \end{itemize}

\item {\bf Institutional review board (IRB) approvals or equivalent for research with human subjects}
    \item[] Question: Does the paper describe potential risks incurred by study participants, whether such risks were disclosed to the subjects, and whether Institutional Review Board (IRB) approvals (or an equivalent approval/review based on the requirements of your country or institution) were obtained?
    \item[] Answer: \answerNA{} %
    \item[] Justification: The paper does not involve crowdsourcing nor research with human subjects.
    \item[] Guidelines:
    \begin{itemize}
        \item The answer NA means that the paper does not involve crowdsourcing nor research with human subjects.
        \item Depending on the country in which research is conducted, IRB approval (or equivalent) may be required for any human subjects research. If you obtained IRB approval, you should clearly state this in the paper. 
        \item We recognize that the procedures for this may vary significantly between institutions and locations, and we expect authors to adhere to the NeurIPS Code of Ethics and the guidelines for their institution. 
        \item For initial submissions, do not include any information that would break anonymity (if applicable), such as the institution conducting the review.
    \end{itemize}

\item {\bf Declaration of LLM usage}
    \item[] Question: Does the paper describe the usage of LLMs if it is an important, original, or non-standard component of the core methods in this research? Note that if the LLM is used only for writing, editing, or formatting purposes and does not impact the core methodology, scientific rigorousness, or originality of the research, declaration is not required.
    \item[] Answer: \answerYes{}%
    \item[] Justification: We clearly described our usage of LLM as a judge, to grade continuations from different models.
    \item[] Guidelines:
    \begin{itemize}
        \item The answer NA means that the core method development in this research does not involve LLMs as any important, original, or non-standard components.
        \item Please refer to our LLM policy (\url{https://neurips.cc/Conferences/2025/LLM}) for what should or should not be described.
    \end{itemize}

\end{enumerate}

\newpage
\appendix

\section{Limitations and Future Work}
\label{app: limitations}
In this work, we investigated three different types of metadata. Yet it is not a very thorough investigation. Furthermore, the results we present in the paper are all from 1.5B models on 100B Fineweb-Edu dataset. It remains to be seen if our findings are consistent when scaling up the model sizes and token counts or switching to different data corpora.

Our work points out a lot of interesting findings from observations in our experiments. The mechanics behind the improved performance brought by URL conditioning are still unclear. From our experimental results, we know it is not because of the indicated topic/format from the URLs, or it could be that the LMs have a different understanding of topic and format domains. Future work should investigate the changes in activations between context-free and context-conditioned pretrained models to have a better understanding.

\section{Broader Impacts}
\label{app: broader-impacts}

As discussed in Section~\ref{sec: context-aware generation}, context-conditioned models are more easily steered. While this capability can be used to guide responses in a desired style or format, it also has the potential to produce harmful or untruthful content. Therefore, these methods should be applied with social responsibility.

\section{More Experimental Details}

Each of the model training run takes around 800 GPU hours on GH200(120GB).

\subsection{Missing figures and tables}

\subsubsection{Acceleration measured by downstream task performances}
\begin{figure}[h!]
    \centering
    \includegraphics[width=\linewidth]{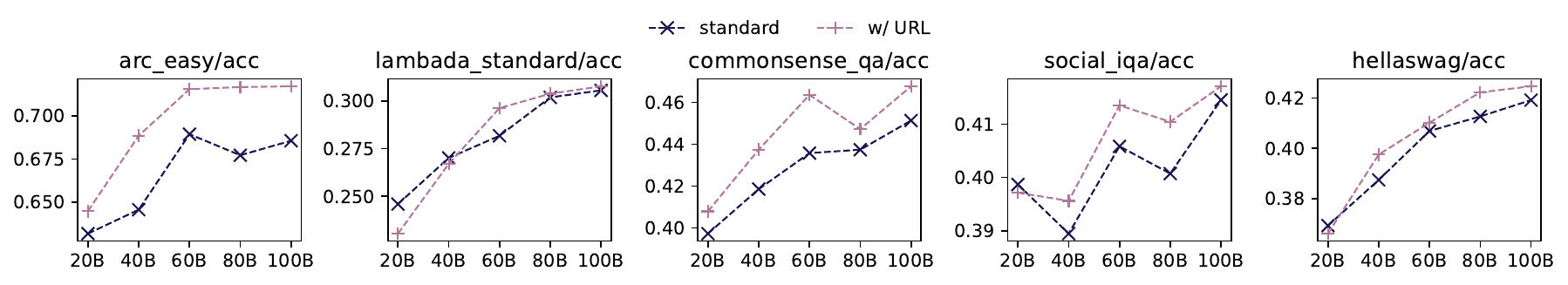}
    \caption{Training acceleration in terms of downstream performances.}
    \label{fig:eval-speed-up-02}
\end{figure}

In addition to Figure~\ref{fig:training-speed-up-eval}, we provide the evaluation with respect to the remaining tasks in Figure~\ref{fig:eval-speed-up-02}. The acceleration is consistent across different tasks.

\subsubsection{Attention pattern to different parts of contexts}
\label{app: attention-pattern}

We randomly sample 100 documents from FineWeb-Edu and prepend each with its corresponding URL, quality score, and domain information. These prepared documents are then input to the corresponding context-conditioned pretrained checkpoints; i.e., URL-prepended documents are fed to the URL-conditioned model. We use a batch size of 1 and a sequence length of 4096, without any padding. The attention weights shown in Figure~\ref{fig: average-attention-pattern-3-subplots} are averaged across all attention heads in the layer and over all non-contextual tokens, and are normalized by the total attention assigned to the contextual portion. 

\begin{figure}
    \centering
    \includegraphics[width=0.7\linewidth]{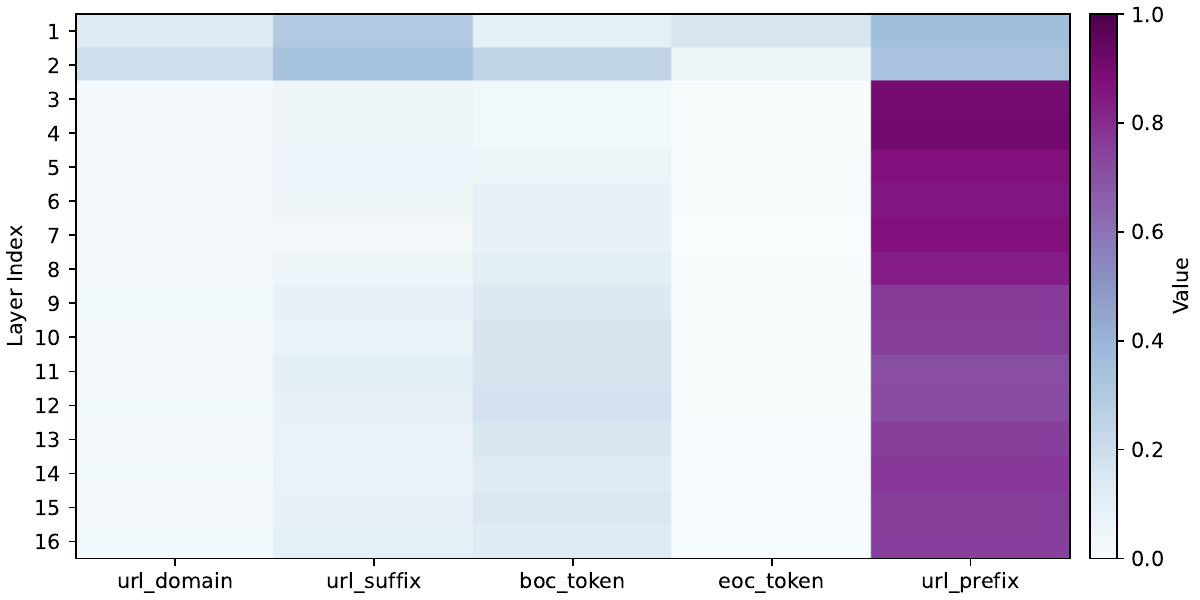}
    \caption{Average attention pattern of URL-conditioned model across layers to different parts of URL context.}
    \label{fig: attention-pattern-heatmap}
\end{figure}

We further visualize the evolution of attention patterns across layers using a heatmap, as shown in Figure~\ref{fig: attention-pattern-heatmap}. Starting from layer 3, we observe a pronounced attention sink on semantically uninformative and commonly shared tokens, such as the URL prefix (\texttt{https://}). This pattern persists across other metadata-conditioned checkpoints. We therefore hypothesize that the attention in the first two layers contributes meaningfully to training acceleration.

\subsubsection{More steering results}

We further present the steering results on the DI-conditioned model in Table~\ref{tab:guidance-with-different-topics} and Table~\ref{tab:guidance-with-different-formats}. Given the exact same prompt, the continuation can be steered in markedly different directions depending on the contextual guidance. Topic-based guidance is highly visible, while format-based guidance is more subtle. For example, when steering with "News Article", the output tends to include concrete details about the composer and adopts a more objective tone—different from the more casual and subjective tone of "Personal Blog". Similarly, when guided by "Spam/Ads", the output adopts a promotional style, often highlighting a specific fair in an advertising manner.

{\ttfamily

\begin{table}[t!]
\caption{Example of different continuations given a prompt conditioned on different topics, while the format domain is kept as \texttt{Knowledge Article}. The continuations are generated in a context-guided fashion using our pretrained checkpoints. (Caution of Hallucination!)}
    \centering
\resizebox{\textwidth}{!}{
\begin{tabular}{p{0.2\linewidth} p{0.8\linewidth}}
\toprule
& \emph{Given prompt}: Paris is known for \\
\toprule
Crime \& Law & its cultural diversity, and LGBTQ+ people, women, and people of color are often underrepresented. The city is home to many LGBTQ+ neighborhoods, which are underrepresented in many other cities and towns. However, these neighborhoods are particularly vulnerable to violence, abuse, and harassment. \\
\midrule
Finance \& Business & its economic prosperity. In the late 19th and early 20th centuries, Paris was a hotbed of the industrial revolution in Europe, and it continues to be a hub for commerce. The city’s robust economy is largely fueled by its high-tech sector, which includes both the manufacturing and technology industries ...
 \\
\midrule
Entertainment &  its rich musical traditions, with a thriving musical scene that has left an indelible mark on the cultural landscape. From the classical era to the modern era, Paris has been a city of musicians and composers who have shaped the music of the world.
 \\
 \midrule 
 Education \& Jobs & its prestigious education system. Universities are renowned for their research capabilities, cutting-edge facilities, and excellent faculty.
To ensure that universities meet the highest standards of academic excellence, France has implemented various quality assurance mechanisms, including external examinations. These examinations assess the overall quality of a university’s academic programs, faculty, and student body.
\\
\bottomrule

    \end{tabular}
    }
    \label{tab:guidance-with-different-topics}
\end{table}
}

{\ttfamily

\begin{table}[t!]
\caption{Example of different continuations given a prompt conditioned on different formats, while the topic domain is fixed as \texttt{Entertainment}. The continuations are generated in a context-guided fashion using our pretrained checkpoints. (Caution of Hallucination!)}
    \centering
\resizebox{\textwidth}{!}{
\begin{tabular}{p{0.2\linewidth} p{0.8\linewidth}}
\toprule
& \emph{Given prompt}: Paris is known for \\
\toprule
News Article & its rich music history, and the city has produced many renowned composers who have left their mark on the world of music. One such composer is Jules Massenet, a French composer who lived from 1842 to 1912. Massenet is considered one of the most important composers of French opera... \\
\midrule
Personal Blog &  its music scene and it is home to some of the most famous musicians in the world. One of the most famous French musicians is Charles Dutoit, who is widely considered to be one of the best French musicians of all time. He has a unique style that blends traditional French music with a modern approach, and he...
 \\
\midrule
Spam/Ads &  its rich musical traditionits many wonderful fairs, particularly the fairs at the Les Demoiselles d'Avignon, but also at the many other fairs throughout the year.
In music, the term "chanson" is often used, especially in the 18th century, to refer to a genre of music that was particularly
\\
\bottomrule

    \end{tabular}
    }
    \label{tab:guidance-with-different-formats}
\end{table}
}

\subsection{Prepended context in inference time}

\newpage
{\ttfamily
\begin{table}[h!]
\caption{Different contexts for Table~\ref{tab: with-or-without-context-lm-eval}. }
    \centering
   \begin{tabular}{p{1.5cm} p{4.5cm} p{4cm} p{2.5cm}}
   \toprule
   Task & URL context & DI context & QS context \\
   \toprule
   Arc-E/C & \url{www.factmonster.com/} & Topic: Science \& Tech., Format: Knowledge Article & \multirow{8}{*}{Quality Score: 5}\\
   CSQA & \url{www.factmonster.com/} & Topic: Social Life, Format: Knowledge Article & \\
   MMLU & \url{www.reference.com/} & Topic: Science \& Tech., Format: Knowledge Article & \\
   PIQA & \url{www.howstuffworks.com/} & Topic: Social Life, Format: Knowledge Article & \\
   SIQA & \url{www.reddit.com/r/AskSocialScience/} & Topic: Social Life, Format: Knowledge Article & \\
   HS & \url{www.wikihow.com/} & Topic: Home \& Hobbies, Format: Knowledge Article & \\
   LBD & \url{www.wikihow.com/} & Topic: Literature, Format: Creative Writing & \\
   WG & \url{www.reddit.com/r/explainlikeimfive/} & Topic: Social Life, Format: Personal Blog & \\
   \bottomrule
   \end{tabular}
   \label{contexts-for-inference}
\end{table}

}

\subsection{LLM Judge}
We use \texttt{gpt-4o} from OpenAI API to evaluate continuations given a prompt. The prompt used for scoring medical \& health domain continuations is presented in the following block. For the history domain, we modify it correspondingly to suit the topic.

\begin{myblock}
    You will be given a prompt and a completion that attempts to continue the prompt.
    Your task is to judge the quality of the completion based on the following criteria:
\newline

    Coherence: Does the continuation flow logically and grammatically from the prompt?
\newline

    Correctness: Is the information medically or scientifically accurate?
\newline

    Reasonableness: Is the claim plausible and not misleading or exaggerated?
\newline

    Relevance: Does it directly relate to and complete the idea expressed in the prompt?
\newline

    Rate the continuation on a scale from 1 to 10, where:
\newline

    1 = incoherent, irrelevant, or medically incorrect
\newline

    5 = somewhat relevant or plausible, but lacking clarity or precision
\newline

    10 = fully coherent, factually accurate, and medically relevant
\newline

    You should ignore the irrelevant answer if the relevant part makes good continuation. 
    You must not hallucinate or invent facts. Use only information in the prompt and completion.
 \newline
 
    Respond in this format:
\newline

    Prompt: {question}
    \newline
    
    Continuation: {answer}
    \newline
    
    Rating: (your rating, as a float between 1 and 10)
    \newline
    
    Rationale: (Brief explanation of the score)
\end{myblock}

\end{document}